\documentclass[letterpaper]{article}
\usepackage[preprint]{aaai2027}
\usepackage[hyphens]{url}
\usepackage{graphicx}
\usepackage{natbib}
\usepackage{caption}
\usepackage{amsmath}
\usepackage{amssymb}
\usepackage{booktabs}
\usepackage{multirow}
\usepackage{array}
\usepackage[table]{xcolor}
\usepackage{tikz}
\usepackage{placeins}
\usepackage{float}

\newcommand{\benchmarkname}{LANTERN}
\newcommand{\xmark}{\(\times\)}
\newcommand{\tabrowrule}{%
  \arrayrulecolor{black!22}\midrule\arrayrulecolor{black}%
}

\title{LANTERN: A Closed-Loop Benchmark for VLM-Based Cooperative Driving with Temporally Grounded Warnings}
\author{
Yongshuo Liu\textsuperscript{\rm 1},
Xu Gao\textsuperscript{\rm 1},
Morui Zhu\textsuperscript{\rm 1},
Yongqi Zhu\textsuperscript{\rm 1},\\
Qi Chen\textsuperscript{\rm 2},
Deyuan Qu\textsuperscript{\rm 2},
Song Fu\textsuperscript{\rm 1},
Qing Yang\textsuperscript{\rm 1}
}
\affiliations{
\textsuperscript{\rm 1}Department of Computer Science and Engineering, University of North Texas, Denton, TX, USA\\
\textsuperscript{\rm 2}Toyota Motor North America, InfoTech Labs, USA
}

\begin{document}
\maketitle

\begin{abstract}
We present \benchmarkname, a closed-loop benchmark for temporally grounded cooperative warnings. \benchmarkname{} separates warning onset, hazard onset, warning termination, and post-hazard recovery, and evaluates each physical event under matched \textit{warning} and \textit{no-warning} executions so that the warning's contribution is measured in isolation rather than confounded with onboard vision. The benchmark spans six safety-critical scenario families and provides 3,272 sequences with 236,309 frames for training, together with 120 matched route pairs for closed-loop evaluation. Each hazard route is evaluated under the \textit{warning} and \textit{no-warning} conditions, while its no-hazard control penalizes unconditional braking. We further introduce the Cooperative Unified Score (CUS), a safety-gated metric that jointly rewards route progress, anticipation, clearance, and recovery. Fine-tuning a representative VLM driving model raises CUS from 34.6 without warnings to 75.5 with them, demonstrating both the value of cooperative warnings and the discriminative power of the paired protocol. All resources will be made publicly available.
\end{abstract}

\section{Introduction}
\begin{figure*}[!t]
    \centering
    \includegraphics[width=0.985\textwidth]{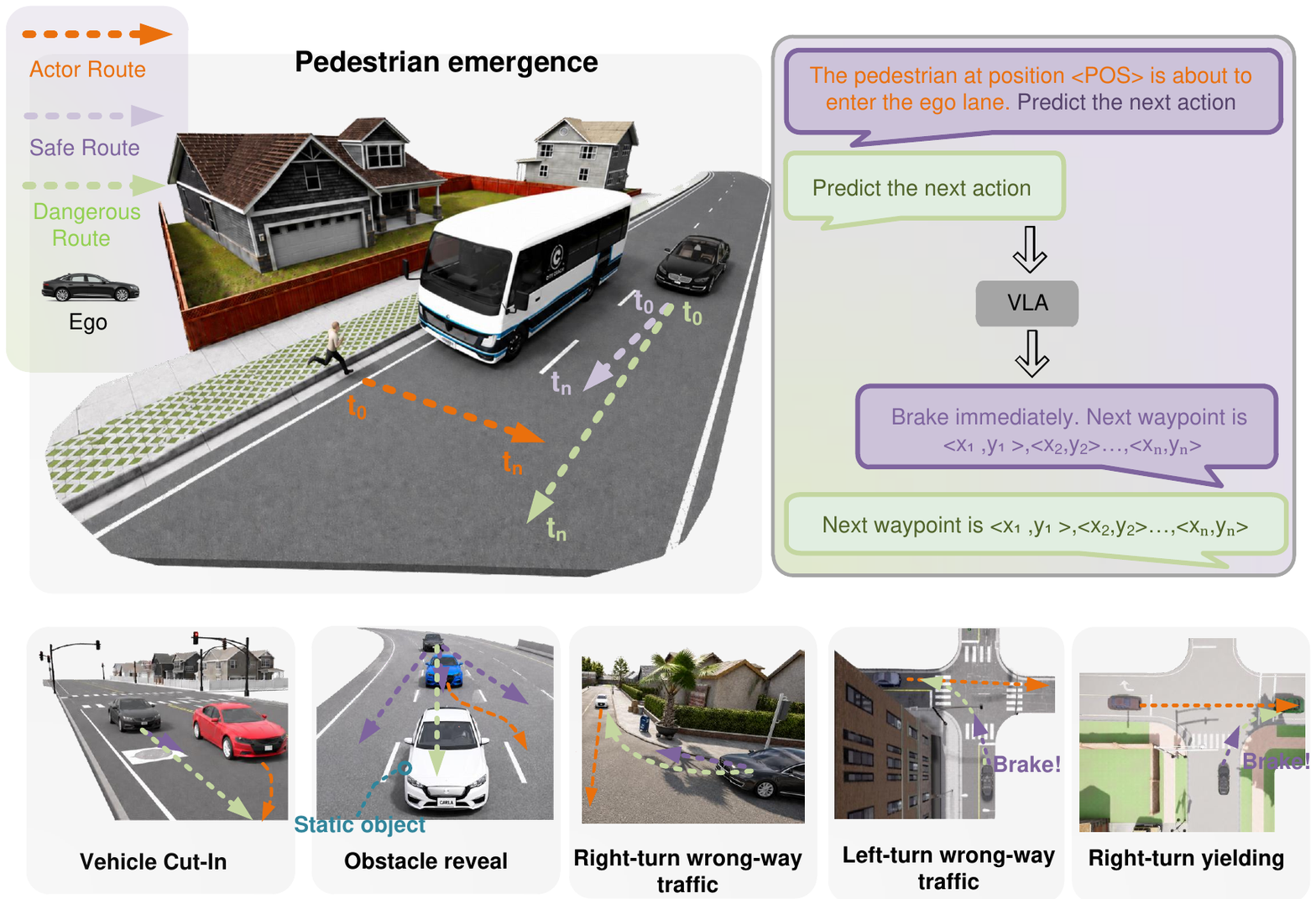}
    \caption{Driving inputs and outputs under the \textit{warning} and \textit{no-warning} conditions. Both conditions receive visual observations, ego state, and navigation state; the \textit{warning} condition additionally receives the cooperative warning during the warning interval.}
    \label{fig:finetuned_simlingo_io_scenarios}
\end{figure*}

Recent advances in vision-language models (VLMs) have enabled
vision-language-action (VLA) driving models that jointly process visual
observations, navigation state, and language for end-to-end driving.
Their language interface provides a natural channel for cooperative
warnings from connected vehicles, roadside units, or traffic-management
services. We define a cooperative warning as an externally supplied
natural-language description of an impending hazard. Such a warning can
describe the hazard before the
relevant evidence is observed through the ego vehicle's onboard sensors.
For example, a roadside unit may report a pedestrian hidden behind an
occluder early enough for the ego vehicle to decelerate safely. In most
cooperative systems, however, shared images, features, intent, or text
are integrated as part of the overall model or communication
configuration. A performance difference can therefore establish that
cooperation helps, but does not by itself attribute the improvement to a
specific warning delivered at a particular stage of a physical event.

Recent work establishes important foundations for language-based and
cooperative driving, but addresses a different evaluation question.
Bench2Drive-VL~\cite{Jia2026bench2drivevl} evaluates general VLM
reasoning and control in closed-loop CARLA scenarios rather than the
effect of an external warning delivered at a specific stage of an
event. V2X-VLM~\cite{You2024v2xvlm} integrates vehicle- and
infrastructure-side observations with textual scene descriptions, so
its results characterize the combined cooperative planning system
rather than the contribution of a single timed warning. Similarly,
MDrive~\cite{Coscoy2026mdrive} evaluates shared perception and
negotiation through complete single- and multi-agent systems, while
LangCoop~\cite{Gao2025langcoop} studies natural-language communication
in terms of collaborative performance and communication efficiency.
Together, these studies evaluate whether cooperation improves an
integrated driving system, but do not isolate a timed warning as a
controlled variable under matched executions of the same evolving event.
Closed-loop evaluation is essential because a warning changes the ego
vehicle's actions and therefore its subsequent observations, interactions,
and recovery.

To address this gap, we introduce \benchmarkname{}, a closed-loop
benchmark built in CARLA~\cite{Dosovitskiy2017carla} that controls whether
and when a cooperative warning is provided. For each safety-critical
scenario, \benchmarkname{} evaluates the same model checkpoint under
matched executions that preserve the route and physical event. We refer
to the executions with and without the warning as the \textit{warning} and
\textit{no-warning} conditions. The benchmark explicitly controls warning onset,
physical-event onset, and warning termination, and ensures that the
event still occurs if the ego vehicle slows or stops before it would
normally begin. \benchmarkname{} covers six scenario families:
Pedestrian Emergence, Vehicle Cut-In, Obstacle Reveal, Left-Turn
Wrong-Way, Right-Turn Wrong-Way, and Right-Turn Yield. These matched
closed-loop executions evaluate whether cooperative warnings induce
earlier and safer responses and whether the ego vehicle resumes progress
after the hazard has passed. The benchmark provides 3,272 training
sequences with 236,309 frames and a closed-loop suite of 120 matched
hazard/control route pairs. The no-hazard controls penalize models that
stop unconditionally in the absence of a physical hazard.

Our contributions are:
\begin{itemize}
    \item We introduce \benchmarkname{}, a closed-loop benchmark comprising six safety-critical scenario families in which hazards are occluded, become visible late, or are observable from outside the ego vehicle.
    \item We develop a temporally controlled, event-matched protocol that aligns warning onset, physical-event onset, and warning termination, making the effect of a cooperative warning independently measurable through anticipation, safety, and post-hazard recovery.
    \item We release an open-loop dataset for learning warning-conditioned behavior and a fixed closed-loop evaluation suite, with aligned front-view and six-camera observation formats for reproducible comparison across sensor interfaces.
    
   \item We introduce the Cooperative Unified Score (CUS), which jointly evaluates safety, route progress, anticipation, recovery, and ordinary driving without a physical hazard. On \benchmarkname{}, cooperative warnings raise the CUS of the same fine-tuned VLA driving model from 34.6 to 75.5, an absolute gain of 40.9 points.
   
\end{itemize}

\section{Related Work}

\begin{table*}[t]
\centering
\scriptsize
\setlength{\tabcolsep}{2.2pt}
\renewcommand{\arraystretch}{1.00}
\begin{tabular*}{\textwidth}{
@{\extracolsep{\fill}}
>{\centering\arraybackslash}m{3.50cm}|
>{\centering\arraybackslash}m{1.20cm}|
>{\centering\arraybackslash}m{0.80cm}|
>{\centering\arraybackslash}m{1.65cm}|
>{\centering\arraybackslash}m{1.05cm}|
>{\centering\arraybackslash}m{0.95cm}|
>{\centering\arraybackslash}m{2.20cm}|
>{\centering\arraybackslash}m{1.45cm}|
>{\centering\arraybackslash}m{2.20cm}
@{}}
\hline\hline
Benchmark
& \multicolumn{5}{c|}{Benchmark Setting}
& \multicolumn{3}{c}{Cooperative Evaluation} \\
\hline
& Domain
& Loop
& Task
& \shortstack{Hazard\\scenarios}
& \shortstack{Training\\data}
& \shortstack{Cooperative\\input}
& \shortstack{Information\\timing}
& \shortstack{Main\\metrics} \\
\hline\hline

\mbox{Bench2Drive \citep{Jia2024bench2drive}}
& CARLA
& Closed
& E2E driving
& \checkmark
& \checkmark
& --
& --
& Driving score, success \\
\arrayrulecolor{black!22}\hline\arrayrulecolor{black}

\mbox{SafeBench \citep{Xu2022safebench}}
& CARLA
& Closed
& Safety testing
& \checkmark
& \xmark
& --
& --
& Safety score \\
\arrayrulecolor{black!22}\hline\arrayrulecolor{black}

\mbox{Bench2Drive-VL \citep{Jia2026bench2drivevl}}
& CARLA
& Closed
& VLM driving
& \checkmark
& \checkmark
& --
& --
& Question answering, score \\
\arrayrulecolor{black!22}\hline\arrayrulecolor{black}

\mbox{V2X-VLM \citep{You2024v2xvlm}}
& Real world
& Open
& Coop. planning
& \xmark
& \checkmark
& Images and text
& Frame-aligned
& $L_2$ error, collision rate \\
\arrayrulecolor{black!22}\hline\arrayrulecolor{black}

\mbox{MDrive \citep{Coscoy2026mdrive}}
& \shortstack{CARLA/\\Real2Sim}
& Closed
& Multi-agent
& \checkmark
& \xmark
& \shortstack{Perception and\\negotiation}
& Online
& \shortstack{Driving score, route\\completion, coop. gain} \\
\arrayrulecolor{black!22}\hline\arrayrulecolor{black}

\mbox{LangCoop \citep{Gao2025langcoop}}
& CARLA
& Closed
& Multi-agent
& \xmark
& \xmark
& Language messages
& Online
& \shortstack{Driving score, route\\completion, bandwidth} \\

\hline\hline

\textbf{\benchmarkname{}}
& CARLA
& Closed
& Coop. VLM driving
& \checkmark
& \checkmark
& Language warnings
& Event-aligned
& \shortstack{CUS, anticipation,\\recovery} \\

\hline\hline
\end{tabular*}

\caption{Comparison of representative driving benchmarks across
benchmark settings and cooperative evaluation protocols. Checkmarks
indicate explicit support, while crosses indicate absence.}
\label{tab:benchmark_comparison}
\end{table*}

\noindent\textbf{Closed-Loop and Safety-Critical Driving Benchmarks.}
CARLA~\cite{Dosovitskiy2017carla} and the CARLA
Leaderboard~\cite{CARLALeaderboard} established simulator-based
evaluation of route completion and driving infractions.
Bench2Drive~\cite{Jia2024bench2drive} broadened this setting to multiple
end-to-end driving capabilities, while SafeBench~\cite{Xu2022safebench}
focused on adversarial and safety-critical traffic conditions.
Waymax~\cite{Gulino2023waymax} and ScenarioNet~\cite{Li2023scenarionet}
support large-scale traffic-scene simulation, whereas
NAVSIM~\cite{Dauner2024navsim} evaluates planning through non-reactive
simulation. AdvSim~\cite{Wang2021advsim} generates physically plausible
actor perturbations, and Fail2Drive~\cite{Gerstenecker2026fail2drive}
uses paired routes to measure generalization under controlled
distribution shifts. These benchmarks establish interactive,
safety-oriented, and paired evaluation, but the controlled variable is
the route, traffic condition, or distribution shift. They do not expose
an externally supplied language warning as an independently scheduled
intervention while holding the driving model and physical hazard fixed.

\noindent\textbf{Language-Conditioned Closed-Loop Driving.}
LMDrive~\cite{Shao2023lmdrive} introduced natural-language navigation
and notice instructions for closed-loop driving, and
SimLingo~\cite{Renz2025cvpr} aligned vision-language understanding with
camera-only driving actions. Drive4C~\cite{Sohn2025drive4c} decomposes
language-guided driving into semantic, spatial, temporal, and physical
capabilities, whereas Bench2ADVLM~\cite{Zhang2025bench2advlm} and
Bench2Drive-VL~\cite{Jia2026bench2drivevl} evaluate more general VLM
reasoning and control in closed-loop environments. DriveLM
~\cite{Sima2023drivelm}, Reason2Drive~\cite{Nie2024reason2drive}, and
STSBench~\cite{FruhwirthReisinger2025stsbench} instead emphasize
question answering, structured reasoning, and spatio-temporal traffic
understanding. ICR-Drive~\cite{Hamid2026icrdrive} further evaluates
sensitivity to paraphrased, ambiguous, noisy, and misleading
instructions under matched simulator settings. In these works, language
is primarily an instruction, query, reasoning target, or persistent
model input. LANTERN instead treats a cooperative hazard warning as a
temporary intervention aligned with the evolution of a physical event,
so that its effect on anticipation and post-warning recovery can be
measured separately.

\noindent\textbf{Cooperative Driving and Language Communication.}
Cooperative driving methods use information from other vehicles or
roadside infrastructure to improve planning beyond the ego vehicle's
onboard observations. OPV2V~\cite{Xu2021opv2v} and
V2X-Sim~\cite{Li2022v2xsim} provide simulated multi-agent perception
benchmarks, while DAIR-V2X~\cite{Yu2022dairv2x} and
TUMTraf-V2X~\cite{Zimmer2024tumtrafv2x} capture real
vehicle--infrastructure settings. Where2comm~\cite{Hu2022where2comm} and
CoBEVT~\cite{Xu2023cobevt} study communication-efficient feature sharing
and cooperative BEV perception. Moving toward planning,
UniV2X~\cite{Yu2025univ2x} integrates vehicle and infrastructure
representations into an end-to-end system,
M3CAD~\cite{Zhu2026m3cad} extends cooperative benchmarking across
perception, tracking, mapping, forecasting, occupancy, and planning,
V2X-VLM~\cite{You2024v2xvlm} combines multimodal V2X observations and
textual scene descriptions for trajectory planning, and
V2X-UniPool~\cite{Luo2025v2xunipool} organizes shared information as a
time-indexed language knowledge pool. V2X-QA~\cite{You2026v2xqa}
evaluates reasoning across ego, infrastructure, and cooperative views
rather than closed-loop control.

Natural-language cooperation is most directly studied by
LangCoop~\cite{Gao2025langcoop}, which compares collaborative and
non-collaborative configurations while emphasizing communication
efficiency and heterogeneous agents. MDrive~\cite{Coscoy2026mdrive}
similarly compares multi-agent and single-agent systems in a broad
closed-loop cooperative benchmark. V2X-VLM also ablates infrastructure
fusion and textual scene prompting. These comparisons establish that
cooperation can improve perception and planning; however, they evaluate
the contribution of a system component or collaborative configuration,
rather than standardizing an individual warning as an event-level
experimental variable with explicit activation and termination.

The distinction is therefore one of evaluation granularity, not whether
prior systems can disable cooperative input.
Table~\ref{tab:benchmark_comparison} situates these benchmarks by their
evaluation setting, cooperative interface, information timing, and
reported metrics. Existing benchmarks principally measure driving
ability, safety, cooperative planning, or communication efficiency at
the system level.
\benchmarkname{} instead controls the warning onset, physical-event
onset, and warning termination within each rollout. This temporal
structure makes anticipation, safety during the event, and recovery
after the warning directly observable, turning the warning itself into
a closed-loop evaluation variable rather than another component of the
cooperative system.

\section{The LANTERN Benchmark}

\subsection{Task Definition and Benchmark Overview}

\noindent\textbf{Driving task and warning interface.}
\benchmarkname{} evaluates an autonomous driving model $f_{\theta}$ in
closed loop. At time $t$, the model receives visual observations $I_t$,
ego state $s_t$, navigation state $r_t$, and an optional cooperative
warning $c_t$, and predicts an $H$-step future ego trajectory,
\begin{equation}
    \hat{\mathbf{W}}_{t+1:t+H}=f_{\theta}(I_t,s_t,r_t,c_t),
\end{equation}
where $c_t=\varnothing$ when no warning is active. The benchmark rolls
out these predictions in CARLA and measures the resulting safety, route
progress, and recovery. Each physical event is executed with the same
driving model and configuration under the \textit{warning} and \textit{no-warning}
conditions.

The warning is an externally supplied benchmark input; hazard detection
and message transmission are outside the evaluation scope. During
simulation, the benchmark controller constructs $c_t$ from the
ground-truth physical event without assuming a particular sender. In
deployment, an equivalent warning could originate from a connected
vehicle, roadside unit, or traffic-management service.

Table~\ref{tab:warning_templates} lists the fixed warning templates. The
token \texttt{\textless{}FRONT\_CAR\_POS\textgreater{}} is not presented
to the model as literal placeholder text. At each step it is grounded by
the relevant hazard actor or conflict location, represented as $(x,y)$ in
meters in the current ego frame, with $x$ forward and $y$ to the right.
The localization diagnostic in Section~\ref{sec:ablations} perturbs this
same grounded position while leaving the warning wording unchanged.

\begin{table}[t]
\centering
\scriptsize
\setlength{\tabcolsep}{4pt}
\renewcommand{\arraystretch}{1.05}
\begin{tabular}{@{}>{\raggedright\arraybackslash}p{0.27\columnwidth}>{\raggedright\arraybackslash}p{0.67\columnwidth}@{}}
\hline\hline
Scenario family & Cooperative warning template \\
\hline\hline
Pedestrian Emergence & The pedestrian at position \texttt{\textless{}FRONT\_CAR\_POS\textgreater{}} is about to enter the ego lane. \\
\tabrowrule
Vehicle Cut-In & The vehicle at position \texttt{\textless{}FRONT\_CAR\_POS\textgreater{}} will cut into the ego lane soon. \\
\tabrowrule
Obstacle Reveal & The static object at position \texttt{\textless{}FRONT\_CAR\_POS\textgreater{}} is blocking the ego lane. \\
\tabrowrule
Left-Turn Wrong-Way & A wrong-way vehicle at position \texttt{\textless{}FRONT\_CAR\_POS\textgreater{}} will cross the ego left-turn path. Yield before turning left. \\
\tabrowrule
Right-Turn Wrong-Way & A wrong-way vehicle at position \texttt{\textless{}FRONT\_CAR\_POS\textgreater{}} will cross the ego right-turn path. Yield before turning right. \\
\tabrowrule
Right-Turn Yield & The vehicle at position \texttt{\textless{}FRONT\_CAR\_POS\textgreater{}} will cross the ego right-turn path. Yield before turning right. \\
\hline\hline
\end{tabular}
\caption{Cooperative warning templates for the six scenario families. Before inference, \texttt{\textless{}FRONT\_CAR\_POS\textgreater{}} is replaced by the hazard or conflict location $(x,y)$ in the current ego frame, measured in meters with $x$ forward and $y$ right.}
\label{tab:warning_templates}
\end{table}

\noindent\textbf{Scenario suite.}
The benchmark comprises six scenario families. In \textbf{Pedestrian
Emergence} (PE), a pedestrian initially hidden by a large roadside vehicle
enters the ego lane. In \textbf{Vehicle Cut-In} (VC), a vehicle traveling in
an adjacent lane abruptly merges into the ego lane. In \textbf{Obstacle Reveal}
(OR), a
leading vehicle changes lane and exposes a stationary object blocking the
ego lane. \textbf{Left-Turn Wrong-Way} (LWW) and \textbf{Right-Turn Wrong-Way}
(RWW)
place an approaching wrong-way vehicle in the ego vehicle's left- and
right-turn paths, respectively. In \textbf{Right-Turn Yield} (RTY), traffic
approaching from the left crosses the ego vehicle's intended right-turn path
and must be yielded to. Together, these scenarios test whether an advance
warning improves responses to hazards that are occluded or not yet directly
observable from the ego vehicle.

\subsection{Closed-Loop Temporal Protocol}
\label{sec:Closed-Loop Temporal Protocol}
\begin{figure}[ht]
    \centering
    \includegraphics[width=\columnwidth]{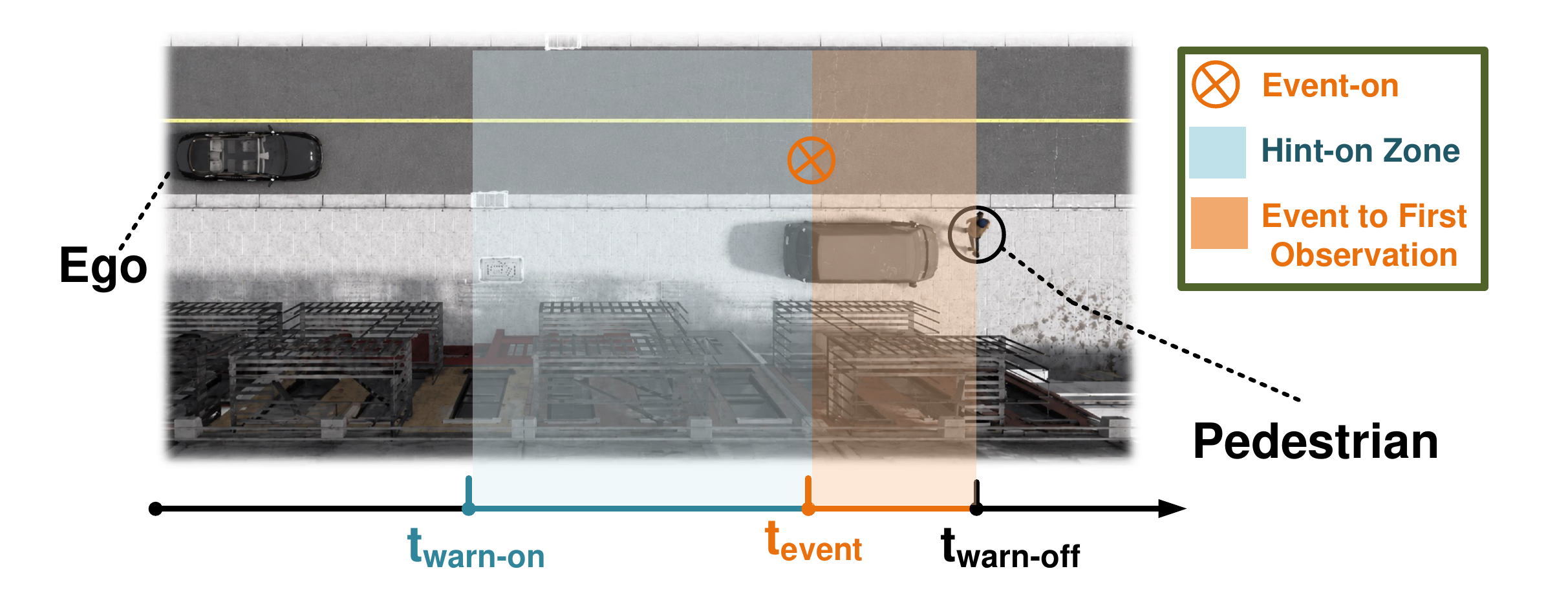}
    \caption{Temporal organization of a \benchmarkname{} scenario. The warning begins before physical-event onset and terminates when the hazard is first directly observed through the ego vehicle's onboard sensors.}
    \label{fig:temporal_protocol}
\end{figure}

\noindent\textbf{Warning and event timing.}
Following Bench2Drive~\cite{Jia2024bench2drive},
each scenario associates a route-relative geometric trigger with a
predefined CARLA actor behavior. Placing the geometric trigger too close to the conflict point can leave the ego vehicle insufficient response time, whereas placing it too far upstream can expose the hazard to onboard sensing before a cooperative warning is necessary.
\benchmarkname{}
therefore places the geometric trigger near the conflict point and activates the warning slightly earlier. As illustrated in
Figure~\ref{fig:temporal_protocol}, we define
\begin{equation}
    t_{\mathrm{warn}}^{\mathrm{on}}
    < t_{\mathrm{event}}
    < t_{\mathrm{warn}}^{\mathrm{off}},
\end{equation}
corresponding to warning onset, physical-event onset, and warning termination.
In CARLA, warning termination is registered after event onset when the hazard
actor enters the $110^\circ$ front-camera view with an unobstructed ray cast.
The \textit{warning} execution receives $c_t=c_{\mathrm{hazard}}$ over
$[t_{\mathrm{warn}}^{\mathrm{on}},t_{\mathrm{warn}}^{\mathrm{off}})$;
the paired \textit{no-warning} execution encounters the same route and physical
event with $c_t=\varnothing$. Recovery is evaluated after the hazard has
passed.

\noindent\textbf{Anti-deadlock rule.}
A warning may cause a driving model to stop before reaching the geometric
trigger and thereby prevent the intended event from occurring. We avoid
this confound by defining
\begin{equation}
    t_{\mathrm{event}}
    =\min\left(t_{\mathrm{geom}},
    t_{\mathrm{warn}}^{\mathrm{on}}+\tau_{\mathrm{force}}\right),
    \qquad \tau_{\mathrm{force}}=2.0~\mathrm{s},
\end{equation}
where $t_{\mathrm{geom}}$ is the time at which the ego vehicle reaches the
geometric trigger. The second term starts a 2-s countdown at warning onset;
if the ego vehicle has not reached the geometric trigger when the countdown
expires, the benchmark activates the physical event. At intersections, the
countdown is paused while a red signal or a traffic queue legitimately
prevents the ego vehicle from proceeding. The rule only affects the scenario actor and never overrides the ego vehicle's control. It is used solely for closed-loop evaluation, whereas open-loop data collection uses the standard geometric trigger.

\subsection{Cooperative Dataset Construction}
\label{sec:dataset_construction}

\noindent\textbf{Collection and A/B controls.}
We use PDM-Lite~\cite{Sima2023drivelm} as the privileged planning expert
adopted by Bench2Drive to execute scenario routes and
provide future-trajectory supervision. Each frame records sensor
observations, ego and navigation states, expert future trajectories,
route and event metadata, and frame-aligned warning annotations. Route A
contains the hazardous behavior and its warning. Route B preserves the
route geometry, navigation objective, and environmental context while
suppressing both, providing ordinary-driving counterexamples that
discourage unconditional braking or yielding.

Open-loop Route A and Route B sequences are generated from corresponding
templates but pass rollout-level quality control independently. They are
therefore complementary training sets rather than guaranteed one-to-one
pairs. This differs from the closed-loop suite below, where every accepted
Route A has one jointly validated Route B counterpart and pair-level
metrics are computed only from that fixed match.

\begin{figure}[t]
    \centering
    \includegraphics[width=\columnwidth]{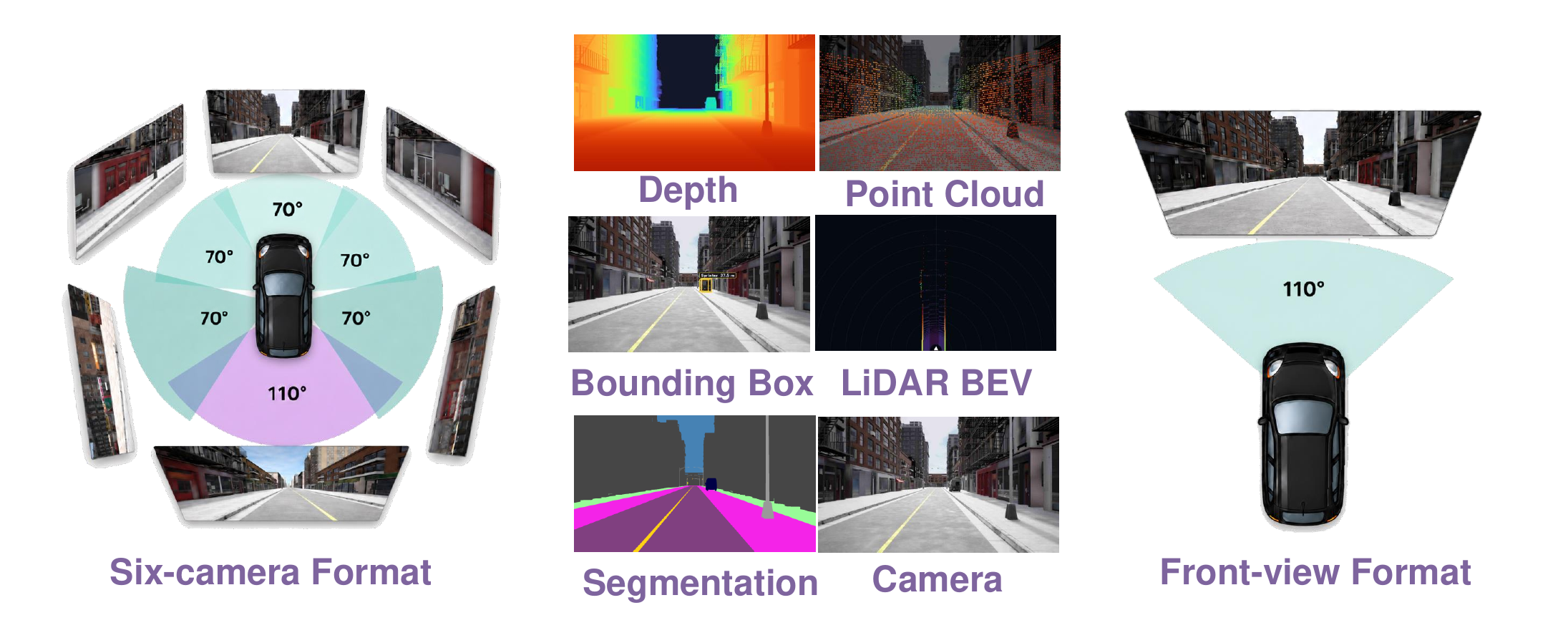}
    \caption{Aligned observation formats and sensor annotations in
    \benchmarkname{}. The six-camera format provides surround coverage, while
    the front-view format provides a single $110^\circ$ camera interface. Both
    support RGB, depth, point clouds, 3D bounding boxes, LiDAR BEV, and semantic
    segmentation.}
    \label{fig:data_formats}
\end{figure}

\noindent\textbf{Observation formats.}
Each accepted sequence is released in two aligned representations: a
six-camera surround-view format and a single-camera front-view format with a
$110^\circ$ horizontal field of view. Both share route and frame identity,
ego and navigation states, event timing, expert trajectories, and warning
annotations. Figure~\ref{fig:data_formats} summarizes the observation layouts
and supported sensor annotations.

\begin{figure}[t]
    \centering
    \includegraphics[width=\columnwidth]{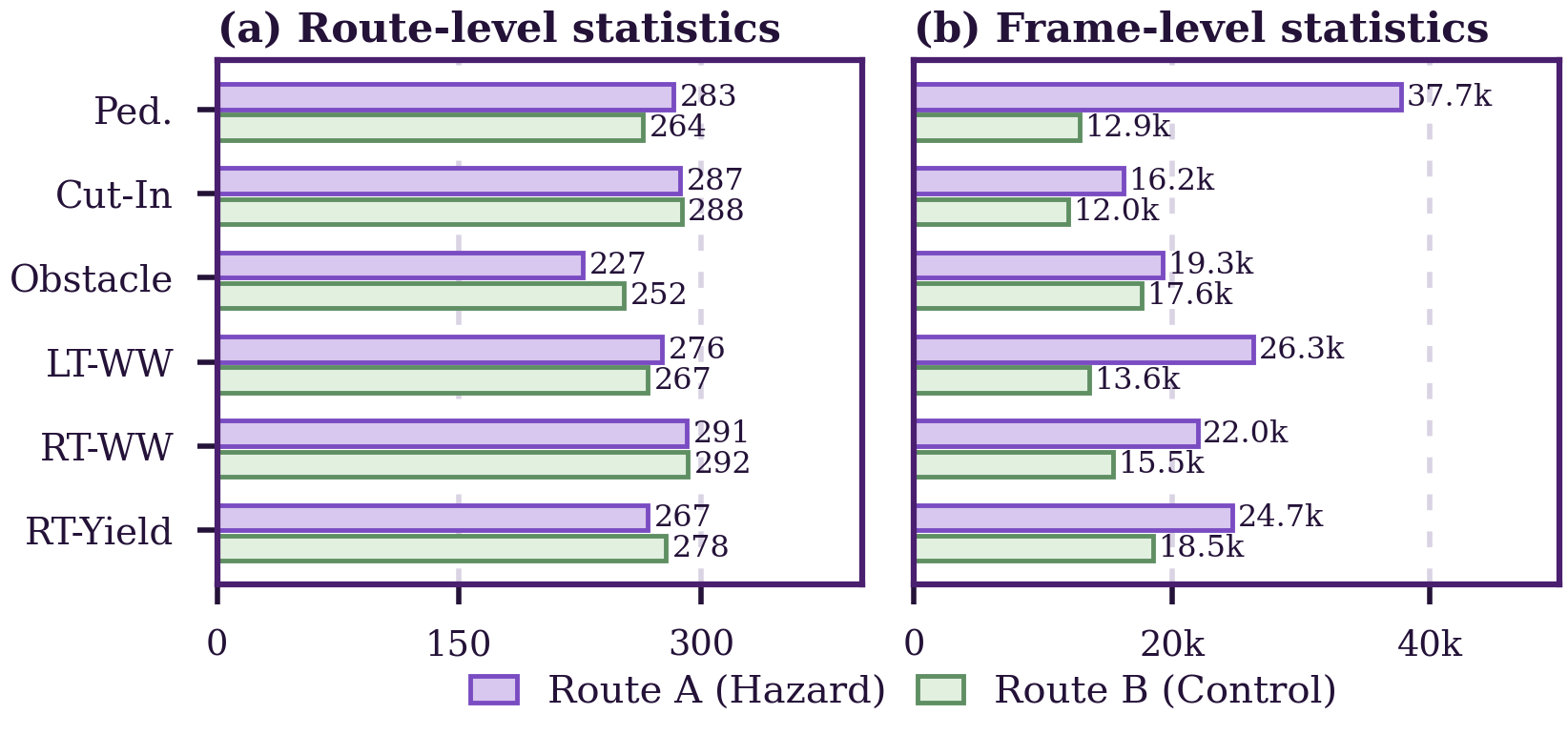}
    \caption{Open-loop dataset composition. Bars report Route A hazard sequences and Route B control sequences at the route and frame levels for each scenario family.}
    \label{fig:openloop_ab_distribution}
\end{figure}

\noindent\textbf{Dataset scale.}
Figure~\ref{fig:openloop_ab_distribution} summarizes 1,631 Route A
sequences with 146,297 frames and 1,641 Route B sequences with 90,012
frames, totaling 3,272 routes and 236,309 frames. The collection spans
cloudy and wet daytime conditions, sunset, clear night, and rainy night.
Independent rollout-level quality filtering accounts for the small Route
A/B count difference. Clips concentrate on hazard approach, interaction,
and recovery, complementing large-scale corpora with dense supervision
for cooperative-warning responses.

\subsection{Closed-Loop Evaluation and Scoring}
\label{sec:closedloop_evaluation}

\noindent\textbf{Evaluation structure.}
The closed-loop suite contains 120 strictly matched Route A/B pairs as defined in Section 3.3, with 20 jointly validated pairs per scenario family. Route A contains the hazard and is executed under both \textit{warning} and \textit{no-warning} conditions: Because the hazard occurs identically in both runs, and the only difference is whether the warning is provided to the ego vehicle, this comparison demonstrates the impact of the cooperative warning on the ego vehicle's control, making it a major contribution of this paper.
It prevents the model from learning to brake based only on visual patterns. Route B has the same scene layout as Route A but contains no hazard. If the model brakes or yields on Route B, it is simply reacting to visual cues instead of a real hazard. This failure is measured by the False Brake-B metric in Table~\ref{tab:cus_decomposition}. Together, the two routes reveal whether the model anticipates hazards from the warning or to the visual scene alone.

\noindent\textbf{Cooperative Unified Score.}
Following the safety-gated aggregation of the Extended Predictive Driver
Model Score~\cite{Cao2025pseudosimulation}, CUS combines multiplicative
safety and validity gates with weighted continuous quality terms. For
matched pair $i$,
\begin{equation}
    \mathrm{CUS}_i
    =100
    \underbrace{\prod_{g\in\mathcal{G}}G_{g,i}}_{\text{hard gates}}
    \underbrace{\sum_{m\in\mathcal{M}}w_m s_{m,i}}_{\text{weighted quality}} .
    \label{eq:cus}
\end{equation}
In Equation~\ref{eq:cus}, $i$ indexes a matched Route A/B pair;
$\mathcal{G}$ is the set of hard-gate criteria, and
$G_{g,i}\in\{0,1\}$ indicates whether pair $i$ passes gate $g$.
$\mathcal{M}$ is the set of continuous quality components,
$s_{m,i}\in[0,1]$ is the normalized score for component $m$, and
$w_m\geq0$ is its weight, with $\sum_{m\in\mathcal{M}}w_m=1$. Thus, any
failed gate sets the pair score to zero; otherwise CUS is the weighted
quality score scaled to 0--100.

The continuous components are Route A Progress (A-Prog.), Route A safety,
Route A temporal quality, and Route B Progress (B-Prog.). A-Prog. and
B-Prog. are the fractions of their respective routes completed. Their
weights are $0.85\times(0.40,0.35,0.25)$ for Route A progress, safety, and
temporal quality, and $0.15$ for Route B progress. Route A safety averages
minimum Time-to-Collision (TTC), normalized at 3~s, and minimum ego--hazard
Clearance (Clear.), normalized at 2~m for pedestrians and static obstacles
and 3~m for vehicle interactions. Temporal quality averages Anticipation
(Ant.) and Recovery (Rec.). Ant. normalizes the lead time of the first
sustained evasive response over the warning-to-event interval: responding
at warning onset scores one, whereas responding at or after event onset, or
not responding, scores zero. Rec. rewards promptly regaining the larger of
2~m/s and half the pre-warning speed while resuming route progress; it
decays linearly to zero at 8~s.

The hard gates reject Route A collision, severe route deviation,
lane-compliance failure, traffic-light or stop-sign violation, blocking or
timeout, and Route B false braking. Gate Pass is the percentage of pairs
satisfying every gate. The result tables report Route A collision
(Coll.-A), route deviation (RD-A), lane-compliance failure (LC-A),
traffic-rule violation (Rule-A), and blocking or timeout (Block-A) rates;
False Brake-B is sustained unjustified stopping in the matched Route B
interval. We report the 20-pair mean for each scenario family, their
equal-weight macro-average, and the components and failure rates underlying
CUS. Operational thresholds are fixed across driving models and released
with the evaluator.

\section{Experiments}

\subsection{Experimental Setup}
The primary evaluation uses the 120 fixed Route A/B pairs defined in
Section~\ref{sec:closedloop_evaluation}, with 20 valid pairs per scenario family
and driving model. Our cooperative baseline, \emph{Coop-SimLingo}, fine-tunes
the original SimLingo checkpoint~\cite{Renz2025cvpr} on \benchmarkname{}
without architectural changes; its action-coupled language interface allows
cooperative warnings to affect driving directly.

We compare Coop-SimLingo under the \textit{warning} and \textit{no-warning}
conditions with original SimLingo as a zero-shot language baseline, the
six-camera VAD planner~\cite{Jiang2023vad}, and the privileged PDM-Lite expert.
Together, they span cooperative and ego-only front-view driving, multi-camera
end-to-end planning, and privileged planning, making this a benchmark
comparison rather than an architecture-only ablation.

The fine-tuning mixture combines hazard-containing Route A samples, hazard-free
Route B controls, and original SimLingo data. Frames whose prediction horizon
crosses warning onset or termination are excluded to preserve temporal label
consistency. Training uses AdamW with a OneCycle schedule for eight epochs,
learning rate $1.5\times10^{-6}$, batch size 4, and validation-loss checkpoint
selection on $2\times$ NVIDIA RTX~A6000 GPUs (approximately 16 hours).
Figure~\ref{fig:finetuned_simlingo_io_scenarios} shows the shared driving
inputs and the additional warning supplied under the \textit{warning} condition.

\subsection{Closed-Loop Results}
\label{sec:closedloop_results}
\begin{table}[!t]
\centering
\scriptsize
\setlength{\tabcolsep}{1.7pt}
\renewcommand{\arraystretch}{1.02}
\begin{tabular*}{\columnwidth}{@{\extracolsep{\fill}}lccccccc@{}}
\hline\hline
Model
& PE
& VC
& OR
& LWW
& RWW
& RTY
& Overall \\
\hline\hline
\rowcolor{purple!12}
\shortstack[l]{Coop-SimLingo\\(\textit{warning})}
& \textbf{88.9} & \textbf{86.5} & \textbf{95.9} & 50.6 & 65.0 & 66.2 & 75.5 \\
\tabrowrule
\rowcolor{green!10}
\shortstack[l]{Coop-SimLingo\\(\textit{no-warning})}
& 8.0 & 59.8 & 55.9 & 4.2 & 17.2 & 62.5 & 34.6 \\
\tabrowrule
\rowcolor{purple!5}
\shortstack[l]{\textit{warning} gain\\\emph{(paired)}}
& \emph{+80.9} & \emph{+26.7} & \emph{+40.0} & \emph{+46.4}
& \emph{+47.8} & \emph{+3.7} & \emph{+40.9} \\
\midrule
\shortstack[l]{SimLingo\\(\textit{warning})}
& 59.0 & 45.4 & 56.3 & 13.8 & 18.2 & 65.6 & 43.1 \\
\tabrowrule
\shortstack[l]{ SimLingo\\(\textit{no-warning})}
& 47.0 & 42.0 & 42.1 & 0.0 & 4.4 & 57.5 & 32.2 \\
\tabrowrule
VAD
& 46.2 & 45.4 & 47.9 & 3.5 & 0.0 & 19.6 & 27.1 \\
\midrule
\shortstack[l]{PDM-Lite\\(Privileged)}
& 81.8 & 83.1 & 95.0 & \textbf{87.2} & \textbf{88.1} & \textbf{88.8} & \textbf{87.3} \\
\hline\hline
\end{tabular*}
\caption{CUS ($\uparrow$) by scenario family. Shading highlights the paired \textit{warning}/\textit{no-warning} comparison and its gain; Overall is the macro-average, and bold marks the best score.}
\label{tab:cus_by_scenario}
\end{table}

\begin{table*}[!t]
\centering
\scriptsize
\setlength{\tabcolsep}{2.0pt}
\renewcommand{\arraystretch}{1.03}
\resizebox{\textwidth}{!}{%
\begin{tabular}{@{}l|ccccccc|cccccc@{}}
\hline\hline
& \multicolumn{7}{c|}{Continuous components and gate pass (\%, $\uparrow$)}
& \multicolumn{6}{c}{Hard-gate failure rate (\%, $\downarrow$)} \\
\cline{2-8}\cline{9-14}
Model
& A-Prog. & TTC & Clear. & Ant. & Rec. & B-Prog. & Gate Pass
& Coll.-A & RD-A & LC-A & Rule-A & Block-A & False Brake-B \\
\hline\hline
Coop-SimLingo (\textit{warning})
& \underline{95.4} & \underline{\textbf{70.3}} & \underline{\textbf{97.4}}
& \underline{86.7} & \underline{\textbf{82.4}} & \underline{93.7}
& \underline{82.5} & \underline{\textbf{0.0}} & \underline{\textbf{0.0}}
& \underline{\textbf{0.8}} & \underline{9.2} & \underline{5.8}
& \underline{1.7} \\
\arrayrulecolor{black!22}\hline\arrayrulecolor{black}
SimLingo (\textit{warning})
& 85.1 & 48.3 & 63.1 & 58.3 & 81.3 & 90.8 & 50.0
& 28.3 & 1.7 & 3.3 & 7.5 & 24.2 & \textbf{0.0} \\
\arrayrulecolor{black!22}\hline\arrayrulecolor{black}
Coop-SimLingo (\textit{no-warning})
& 89.2 & 34.8 & 44.4 & 53.5 & 62.7 & 93.7 & 43.3
& 41.7 & \textbf{0.0} & 1.7 & 8.3 & 19.2 & 1.7 \\
\arrayrulecolor{black!22}\hline\arrayrulecolor{black}
SimLingo (\textit{no-warning})
& 83.7 & 38.4 & 46.4 & 54.6 & 60.1 & 90.8 & 37.5
& 46.7 & 0.8 & 2.5 & 12.5 & 33.3 & \textbf{0.0} \\
\arrayrulecolor{black!22}\hline\arrayrulecolor{black}
VAD
& 68.1 & 44.6 & 53.0 & 49.1 & 62.3 & 75.3 & 35.0
& 28.3 & 0.8 & 9.2 & 6.7 & 42.5 & \textbf{0.0} \\
\hline\hline
PDM-Lite (Privileged)
& \textbf{98.5} & 58.2 & 93.5 & \textbf{93.1} & 78.5 & \textbf{98.5} & \textbf{97.5}
& \textbf{0.0} & \textbf{0.0} & \textbf{0.8} & \textbf{0.0} & \textbf{1.7} & \textbf{0.0} \\
\hline\hline
\end{tabular}%
}
\caption{CUS decomposition over 120 matched Route A/B pairs. Continuous components and Gate Pass are higher-is-better, hard-gate failure rates are lower-is-better, and column-best values are bold.}
\label{tab:cus_decomposition}
\end{table*}

Table~\ref{tab:cus_by_scenario} foregrounds the paired comparison that
isolates the value of cooperative warnings. For the same fine-tuned
checkpoint, enabling the warning raises overall CUS from 34.6 to 75.5, a
40.9-point gain. Original SimLingo gains only 10.9 points when given the
same warnings (32.2 to 43.1). The contrast shows that the language
interface alone provides limited benefit; fine-tuning is what aligns the
warning content and timing with an appropriate driving response.

The scenario pattern explains why this alignment is effective. The gain
is largest for Pedestrian Emergence (+80.9), followed by Right-Turn
Wrong-Way (+47.8), Left-Turn Wrong-Way (+46.4), and Obstacle Reveal
(+40.0). In these families, the relevant road user or obstacle is
occluded, outside the front-camera view, or directly observed only after
the available reaction distance has narrowed. The warning supplies that
missing evidence before onboard observation, extending the interval in
which the model can decelerate or yield. The smaller gain for Right-Turn
Yield (+3.7) is also informative: cross traffic is often already visible
near the turn, and success depends more on gap selection than on early
hazard awareness. Cooperative warnings therefore help most when they add
information that the ego observation does not yet contain, rather than
uniformly inflating performance.

The decomposition in Table~\ref{tab:cus_decomposition} supports this
mechanism. Relative to the \textit{no-warning} execution of the same checkpoint,
the \textit{warning} condition increases TTC from 34.8 to 70.3, clearance from
44.4 to 97.4, anticipation from 53.5 to 86.7, and recovery from 62.7 to
82.4, while reducing Route A collisions from 41.7\% to zero. This is an
earlier and safer response, not merely a more conservative terminal
state: Route B progress remains 93.7 and sustained false braking is only
1.7\%. PDM-Lite remains the strongest overall privileged reference,
especially at intersections, but Coop-SimLingo with warnings is
competitive in the late-visible hazard families despite relying on
camera observations and the cooperative language input.

The remaining failures are concentrated in interaction quality rather
than initial hazard recognition. A warning can create enough time to
avoid an immediate collision without specifying a dynamically feasible,
rule-compliant maneuver; intersection negotiation, excessive caution,
and delayed recovery can therefore still prevent route completion. This
distinction motivates evaluating anticipation, safe interaction, and
recovery jointly instead of treating collision avoidance alone as
success.

\begin{table}[H]
\centering
\scriptsize
\setlength{\tabcolsep}{2.7pt}
\renewcommand{\arraystretch}{1.03}
\begin{tabular*}{\columnwidth}{@{\extracolsep{\fill}}lcccc@{}}
\hline\hline
Scenario
& ADE $\downarrow$
& FDE $\downarrow$
& DAC $\uparrow$
& \shortstack{Warning\\Response $\uparrow$} \\
\hline\hline
Pedestrian Emergence & 2.88 & 5.76 & \textbf{100.0} & 93.0 \\
\tabrowrule
Vehicle Cut-In & \textbf{2.29} & \textbf{4.65} & \textbf{100.0} & 88.0 \\
\tabrowrule
Obstacle Reveal & 2.54 & 4.89 & \textbf{100.0} & 96.0 \\
\tabrowrule
Left-Turn Wrong-Way & 3.03 & 6.62 & 99.0 & \textbf{99.0} \\
\tabrowrule
Right-Turn Wrong-Way & 2.87 & 6.41 & \textbf{100.0} & 98.0 \\
\tabrowrule
Right-Turn Yield & 2.81 & 6.17 & 95.0 & 95.0 \\
\hline
Overall & 2.74 & 5.75 & 99.0 & 94.8 \\
\hline\hline
\end{tabular*}
\caption{Zero-shot open-loop diagnostics on nuScenes using 100 frames per scenario family. Average Displacement Error (ADE) and Final Displacement Error (FDE) are reported in meters; Drivable-Area Compliance (DAC) and Warning Response are percentages.}
\label{tab:nuscenes_realworld}
\end{table}

\subsection{Real-World Transfer}
To examine whether the learned warning-conditioned behavior extends beyond
simulation, we evaluate Coop-SimLingo on nuScenes~\cite{Caesar2020nuscenes}.

Because nuScenes does not annotate cooperative warnings, we select 600
scene-disjoint frames, with 100 frames per scenario family, from contexts in
which the corresponding hazards could plausibly emerge. Three calibrated
forward cameras are reprojected into the model's $110^\circ$ input, and each
frame is evaluated under paired \textit{no-warning} and \textit{warning} conditions. No
nuScenes sample is used for training or model selection.

\begin{figure}[t]
    \centering
    \includegraphics[width=\columnwidth]{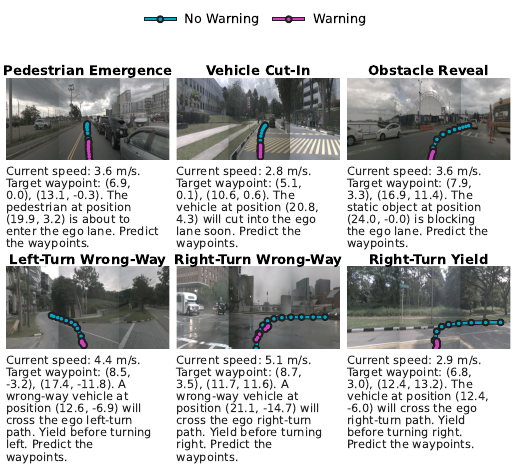}
    \caption{Paired nuScenes examples for Coop-SimLingo. Cyan and magenta
    show \textit{no-warning} and \textit{warning} predictions from the same frame; each panel
    reproduces the complete \textit{warning} input below the image.}
    \label{fig:nuscenes_paired_warning}
\end{figure}

As shown in Table~\ref{tab:nuscenes_realworld}, the \textit{no-warning} predictions
obtain an ADE/FDE of 2.74/5.75~m. These errors are dominated by longitudinal
displacement because Coop-SimLingo predicts more conservative speeds than the
recorded drivers. Its lateral ADE is 1.93~m overall and only 0.15~m on
straight-driving samples, indicating that the predicted paths generally
preserve the correct road geometry. Under warnings, 99.0\% of trajectories
remain within the observed drivable area, and 94.8\% produce a clear
longitudinal or lateral response. The remaining cases mainly involve brief
deceleration or progress-increasing avoidance that falls outside the fixed
response threshold, rather than invalid trajectories. Together with
Figure~\ref{fig:nuscenes_paired_warning}, these results demonstrate that
Coop-SimLingo produces consistent and geometrically plausible
warning-conditioned behavior on real-world images.

\subsection{Ablations and Robustness}
\label{sec:ablations}
Table~\ref{tab:ablation_results} tests perturbations to warning timing,
event activation, localization, and wording against the same Route~B control
and CUS computation. Event Exposure (Event Exp.) is the percentage of
executions in which the intended physical event begins.

\begin{table}[!ht]
\centering
\scriptsize
\setlength{\tabcolsep}{1.7pt}
\renewcommand{\arraystretch}{1.02}
\begin{tabular*}{\columnwidth}{@{\extracolsep{\fill}}lrrrrrrr@{}}
\hline\hline
Variant & \shortstack{CUS\\$\uparrow$} &
\shortstack{A-Prog.\\$\uparrow$} &
\shortstack{Event\\Exp. $\uparrow$} &
\shortstack{Ant.\\$\uparrow$} &
\shortstack{Rec.\\$\uparrow$} &
\shortstack{Coll.-A\\$\downarrow$} &
\shortstack{Block-A\\$\downarrow$} \\
\hline\hline
Full temporal protocol & 75.5 & 95.4 & 99.2 & 86.7 & 82.4 & 0.0 & 5.8 \\
\tabrowrule
Late warning onset & 54.9 & 95.2 & 100.0 & 0.0 & 73.8 & 14.9 & 7.6 \\
\tabrowrule
Warning beyond termination & 63.3 & 79.6 & 100.0 & 84.5 & 64.1 & 0.0 & 29.7 \\
\tabrowrule
No anti-deadlock rule & 51.2 & 79.0 & 71.0 & 87.0 & 74.2 & 9.4 & 35.7 \\
\tabrowrule
Noisy warning position & 75.5 & 98.4 & 100.0 & 87.4 & 82.1 & 3.4 & 3.3 \\
\tabrowrule
Held-out warning paraphrase & 57.7 & 85.9 & 100.0 & 78.6 & 75.8 & 25.7 & 5.9 \\
\hline\hline
\end{tabular*}
\caption{Sensitivity to warning timing, event activation, localization, and wording. Each variant changes one component of the full protocol; all entries use a 0--100 scale.}
\label{tab:ablation_results}
\end{table}

Late warning onset removes the lead time to act, reducing Ant. to zero and
raising collisions to 14.9. Extending the warning 10~s beyond termination
instead delays recovery and raises blocking to 29.7. Without the anti-deadlock
rule, event exposure falls to 71.0 and blocking reaches 35.7, confirming the
need to activate the intended event. Two-meter position noise leaves CUS at
75.5, whereas held-out paraphrases reduce it to 57.7, indicating greater
sensitivity to wording than to moderate localization noise.
\FloatBarrier

\section{Conclusion}
We introduced \benchmarkname, a closed-loop benchmark for temporally controlled and independently measurable cooperative warnings. Its paired protocol evaluates whether warning-conditioned responses are timely, safe, and recoverable. Across the evaluated hazards,
warnings provide the greatest benefit when relevant evidence is occluded
or becomes visible only after the available reaction distance has
narrowed. The results also show that advance information alone is
insufficient: warning timing, interaction planning, and post-hazard
recovery remain important determinants of successful driving. Open-loop
evaluation on nuScenes further indicates that the learned
warning-conditioned behavior persists under real-world visual inputs.
\benchmarkname{} currently relies on structured warnings derived from
simulator state and focuses on individual hazards. Future work will study learned warning sources, communication uncertainty, and interacting hazards.

\bibliography{references}

\clearpage
\onecolumn
\appendix
\section{Data Collection Details}

The main paper summarizes the released data and observation formats. This
section instead documents how route geometry is recovered, materialized, and
accepted into the release.

\noindent\textbf{Route recovery and geometric diversity.}
Routes are recovered from recorded simulator trajectories rather than
constructed from sparse hand-written coordinates. The released suite combines
routes inherited from the original SimLingo collection with an expansion
scanned from Bench2Drive-Base. For the expansion, a CARLA
\texttt{GlobalRoutePlanner} with 1-m sampling resolution reconstructs the
drivable path between recorded control points. Candidate geometry is then
resampled at 2-m arc-length intervals for overlap testing and hashing; actors
and scenario state from the source clip are not copied.

Each candidate must provide a normal approach, a conflict point, and at least
29~m of planned route after the conflict for recovery. Pedestrian Emergence,
Vehicle Cut-In, and Obstacle Reveal use predominantly straight segments.
Turning scenarios must traverse a junction with at least three distinct road
arms, excluding two-arm corner geometries. All scenario markers must project
within 1~m of the planner-reconstructed route and follow the ordering required
by the scenario protocol. In the final closed-loop suite, route lengths are
68.0--134.6~m for Pedestrian Emergence, 58.0--163.8~m for Vehicle Cut-In,
66.0--133.1~m for Obstacle Reveal, 56.0--80.4~m for Left-Turn Wrong-Way,
52.5--87.3~m for Right-Turn Wrong-Way, and 50.4--74.6~m for Right-Turn
Yield.

Every candidate retains its source clip, source route, and geometry hash. A
source clip may appear only once within a scenario family. Two routes in the
same town and family are treated as near duplicates when, in both directions,
at least 70\% of their 2-m samples lie within 3~m of the other route; the second
candidate is rejected. The scanner also checks overlap with the existing
closed-loop suite and open-loop training routes. Consequently, additional
examples are not produced by shifting a sliding window along the same
trajectory or by changing only weather or actor appearance. Geometry may be
reused across different scenario families because the controlled interaction
is different.

\noindent\textbf{Scenario materialization and controls.}
Accepted candidates are ranked jointly by town, source scenario, weather,
route geometry, and junction structure before physical actors and temporal
markers are inserted. Materialization records the route-relative arm, warning,
event, conflict, and recovery locations together with actor parameters and
source provenance. Route A and its Route B control have identical waypoints,
town, weather, navigation objective, route length, and background-traffic
seed. Route B is not an empty road: it retains occluders, candidate vehicles,
and non-conflicting traffic while disabling only the target hazardous behavior
and its warning. For example, the pedestrian occluder remains without a
crossing pedestrian, and the cut-in vehicle remains in its adjacent lane
without merging.

Background traffic remains enabled in both variants. To prevent a foreground
vehicle from becoming an alternative explanation for stopping at an
intersection, the dynamic audit detects persistent same-direction traffic in
the ego route corridor: a background actor is flagged when it remains
1--25~m ahead, within 1.8~m laterally, and aligned with the ego direction for
at least three stored frames. Such a pair is corrected and reevaluated while
adjacent-lane, opposing, and other contextual traffic is retained. Open-loop
Route A and Route B rollouts pass quality control independently; closed-loop
Route A/B files are admitted only as jointly validated pairs.

\noindent\textbf{Expert rollout and supervision.}
PDM-Lite executes the selected routes as the privileged planning expert used
for data collection.
CARLA advances at 20~Hz, and every fifth simulator tick is retained, producing
4-Hz supervision. Each retained state is paired with ten future expert
waypoints at 0.25-s intervals. Waypoints are expressed in meters in the current
ego frame, with $x$ forward and $y$ right. Sensor observations and annotations
are accepted only when their frame indices agree with the measurements,
navigation state, expert trajectory, and warning-event timeline.

\noindent\textbf{Open-loop acceptance.}
An open-loop rollout requires a valid evaluator record, at least 95\% route
completion, no recorded collision, and complete synchronized modalities.
Route A must contain an ordered event trace with at least three retained frames
before warning onset,
$t_{\mathrm{warn}}^{\mathrm{on}} \leq t_{\mathrm{event}} <
t_{\mathrm{warn}}^{\mathrm{off}}$, with equality permitted only for
Right-Turn Yield, an explicit physical-event completion record, and a
downstream recovery segment. Scenario-specific checks reject
failed actor spawning, invalid pose or motion, missing warning text, incomplete
pedestrian crossings, and low-speed tails longer than 40 retained frames.
Route B must contain neither an active warning nor a hazardous actor lifecycle
and must pass the same completion, collision, and synchronization tests.

\noindent\textbf{Closed-loop static validation.}
Before CARLA execution, each XML file must contain exactly one route and one
scenario, with consistent identifiers, family, and town. The validator
reconstructs the route with the CARLA planner and rejects a candidate when its
planned length differs from the recorded candidate length by more than the
larger of 5~m or 10\%. It then checks marker projection and ordering, configured
warning and event lead distances, recovery length, turn direction, and
junction arm count. Pair-level checks require exact Route A/B agreement in
waypoints and weather, valid control flags, retained context parameters, and
unique source and geometry hashes. Static errors must be resolved before a
route is run in CARLA.

\noindent\textbf{Dynamic and visual acceptance.}
PDM-Lite first evaluates three Route A examples from each family as a smoke
gate; the remaining routes are run only after all three pass. A Route A
execution must complete the route, incur no collision, blocking, or scenario
timeout, emit warning onset and termination exactly once, and record the
family-specific physical completion event. A Route B execution must complete
without these infractions, emit no warning or hazard lifecycle, preserve its
context actors, and contain enough telemetry to cover the Route A warning
interval. Every run must additionally provide at least 20 RGB frames and
nonempty videos.

Infrastructure failures, including CARLA crashes, world-loading errors, and
port conflicts, are retried at most three times and are kept separate from
route failures. Reproducible actor, geometry, timing, or navigation failures
require correcting or replacing the candidate and rerunning both variants;
metric files are never edited to convert a failure into a pass. Manual review
then checks the approach, warning onset, event onset, maximum interaction,
event clearance, warning termination, recovery, and the matched Route B
context. After full acceptance, three Route A and three Route B examples per
family are rerun under independent CARLA launches to expose stochastic
background-traffic failures. The final v1.1 release freezes 120 pairs, all
route and scenario hashes, source provenance, validation reports, and
SHA-256 checksums.

\section{Town, Weather, and Time Distribution}

\begin{figure}[htbp]
    \centering
    \includegraphics[width=\textwidth]{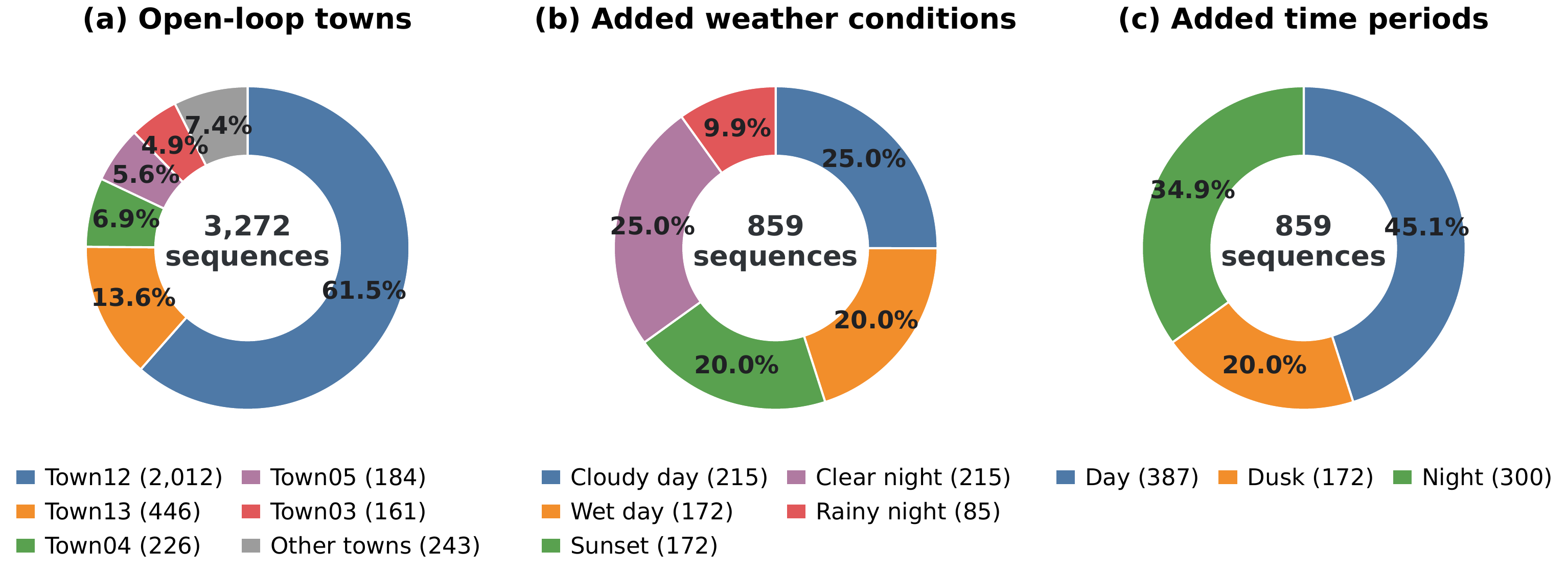}
    \caption{Sequence-level distribution of the open-loop release and its
    environmental extension. (a) Town coverage for all 3,272 sequences.
    (b) Weather profiles and (c) time periods for the 859-sequence extension.}
    \label{fig:supp_town_weather_time_distribution}
\end{figure}

The complete open-loop release contains 3,272 sequences and 236,309 frames:
1,631 Route A sequences with 146,297 frames and 1,641 Route B sequences with
90,012 frames. Figure~\ref{fig:supp_town_weather_time_distribution}(a) shows
the sequence-level geographic distribution. The open-loop pool covers ten
CARLA towns and is concentrated in Town12 because that map provides the
largest number of route segments satisfying the scenario geometry and quality
filters. For readability, ``Other towns'' combines Town01, Town02, Town06,
Town10HD, and Town15; Town07 has no accepted open-loop sequence. The
closed-loop suite is selected for valid, visually distinct interactions rather
than to reproduce this frequency: it covers eleven towns, with 35 pairs in
Town05, 21 in Town04, 19 in Town12, 14 in Town13, and the remaining 31 across
seven towns.

The initial open-loop collection contributes 2,413 sequences and 168,028
frames under its route-defined conditions. A controlled environmental
extension adds 859 accepted sequences and 68,281 frames. As shown in
Figure~\ref{fig:supp_town_weather_time_distribution}(b--c), the extension
contains 215 cloudy-day, 172 wet-day, 172 sunset, 215 clear-night, and 85
rainy-night sequences, corresponding to 387 daytime, 172 dusk, and 300
nighttime sequences. Route A and Route B are sampled from the same candidate
and environment pools, but their final counts differ because quality filtering
is applied independently.

For the extension, cloudiness, rain, wetness, fog, and sun altitude are
$(80,0,0,5,35)$ for cloudy day, $(75,35,70,10,30)$ for wet day,
$(35,0,0,5,8)$ for sunset, $(10,0,0,5,-20)$ for clear night, and
$(90,35,80,15,-15)$ for rainy night. The first four entries use CARLA's
0--100 scale and sun altitude is in degrees.

The closed-loop suite uses five fixed profiles---clear day, cloudy day, wet
day, light rain, and dusk---with 24 matched pairs per profile. Each scenario
family contributes four pairs to every profile. Listing cloud, rain, wetness,
fog, and sun altitude in that order, their CARLA settings are 5, 0, 0, 2, and
45 for clear day; 75, 0, 0, 3, and 55 for cloudy day; 65, 0, 70, 5, and 35 for
wet day; 85, 25, 80, 4, and 40 for light rain; and 40, 0, 10, 2, and 8 for
dusk. The first four values use CARLA's 0--100 scale and sun altitude is in
degrees. Route A and Route B within a closed-loop pair share the same town and
environmental profile exactly. Thus, environmental imbalance can affect
open-loop training frequency but cannot confound a closed-loop pairwise
comparison.
\FloatBarrier

\section{NPC and Scenario-Actor Behavior}

\noindent\textbf{Background traffic.}
Closed-loop evaluation keeps Bench2Drive background activity enabled. CARLA
and Traffic Manager run synchronously with hybrid physics and Traffic Manager
seed 0. The number of background actors is not imposed as a scenario-level
constant: Bench2Drive's route background behavior determines feasible spawn
locations from the current road topology. Ordinary background vehicles remain
Traffic Manager controlled and obey its lane-following, car-following, signal,
and lane-change logic. The same frozen route, weather profile, and evaluator
seed are used across compared conditions.

\noindent\textbf{Scenario actors.}
Hazard actors are separate from background traffic and carry the
\texttt{scenario} role. Their motion is deterministic and implemented by a
scenario-specific behavior tree. The pedestrian follows a fixed crossing at
3~m/s; the cut-in vehicle executes a timed lane change with a minimum speed of
6.5~m/s; the reveal vehicle changes lane to expose a stationary object;
wrong-way vehicles follow the reversed ego route at 7--9~m/s; and Right-Turn
Yield generates a finite cross-traffic passage at 8--12~m/s. A scenario actor
is removed only after its scripted interaction has completed or it has cleared
the conflict region. The cut-in vehicle is instead returned to Traffic Manager
after completing its merge, so it continues as ordinary traffic.

Route B preserves visual context without executing the hazard. Pedestrian
Emergence retains the occluder but removes the pedestrian; Vehicle Cut-In
keeps the vehicle in the adjacent lane under Traffic Manager; Obstacle Reveal
keeps the lead-vehicle context but does not create the blocking object;
wrong-way actors are rotated into the legal direction and handed to Traffic
Manager; and Right-Turn Yield keeps a stationary cross-traffic context actor
without initiating the crossing flow. Route B contains no warning.

\noindent\textbf{Interaction safeguards.}
Background traffic is preserved except where it would invalidate the intended
interaction. Before and during a route, a corridor filter removes only
non-scenario vehicles lying ahead on the ego route, up to the configured
55-m runtime clearance. For turning scenarios, new background spawning on the
ego entry lane is also disabled while traffic on adjacent, opposing, and cross
lanes remains active. Vehicle Cut-In additionally clears a 25-m ego-lane
interaction region, and Obstacle Reveal removes only vehicles within 14~m of
the reveal vehicle's spawn area. Hero and scenario actors are never removed by
these filters. These safeguards prevent an unrelated lead vehicle from causing
the ego vehicle to stop, or from colliding with a hazard actor before the
controlled event, while retaining unrelated traffic needed for realistic
closed-loop interaction.

\section{Detailed Scenario Specifications}

All six families share the same observable lifecycle. Warning onset occurs
when the ego vehicle reaches the route-relative warning point. The physical
event begins when the ego reaches its geometric trigger or, for the first five
families, when the 2-s anti-deadlock interval expires after warning onset. The
Right-Turn Yield event point coincides with its warning point and therefore
requires no forced trigger. Warning termination is registered only after the
event begins, when the controlled hazard (actor or obstacle) enters the ego
vehicle's $110^\circ$ front-camera frustum and at least one ray from the
virtual camera to the hazard is unobstructed. The evaluator then continues
through event clearance and a
downstream recovery segment. Table~\ref{tab:supp_scenario_specifications}
gives the family-specific realization of this lifecycle. Warning and event
lead distances are measured upstream from the route-relative conflict point.

\begin{table}[H]
    \centering
    \small
    \setlength{\tabcolsep}{1.2pt}
    \renewcommand{\arraystretch}{1.16}
    \begin{tabular}{>{\raggedright\arraybackslash}m{2.35cm}
                    >{\raggedright\arraybackslash}m{4.05cm}
                    >{\centering\arraybackslash}m{1.15cm}
                    >{\centering\arraybackslash}m{1.15cm}
                    >{\raggedright\arraybackslash}m{3.25cm}
                    >{\raggedright\arraybackslash}m{3.65cm}}
        \hline\hline
        Scenario & Physical event & Warn lead & Event lead & Actor parameters & Clearance and Route B control \\
        \hline\hline
        Pedestrian Emergence & A pedestrian emerges from behind a large roadside occluder and crosses the ego lane. & 18~m & 9~m & 3~m/s; 22-m crossing; 30-m recovery & Walker and occluder are removed after the crossing. Route B retains only the occluder. \\
        \tabrowrule
        Vehicle Cut-In & A vehicle 21~m ahead in an adjacent same-direction lane merges into the ego lane. & 10/17~m & 4~m & 80--85\% reference speed; $\geq$6.5~m/s; 2-s lane change & The actor is handed to Traffic Manager after the merge. Route B keeps it in the adjacent lane. \\
        \tabrowrule
        Obstacle Reveal & A lead van changes lane, exposing a stationary object in the ego lane. & 12~m & 6~m & Lead at 12--20~km/h; object 12--16~m ahead; 8-s stop limit & The object is removed after the ego response or stop limit. Route B retains the lead vehicle but creates no object. \\
        \tabrowrule
        Left-Turn Wrong-Way & A wrong-way vehicle follows the reverse ego route through the left-turn conflict. & 10~m & 4~m & 7--9~m/s; 25-m spawn offset; 15-m clearance target & Actor is removed after passing the conflict and traveling at least 5~m. Route B sends it in the legal direction. \\
        \tabrowrule
        Right-Turn Wrong-Way & A wrong-way vehicle follows the reverse ego route through the right-turn conflict. & 10~m & 4~m & 7--9~m/s; 20-m spawn offset; 20-m clearance target & Actor is removed after clearing the turn path. Route B sends it in the legal direction. \\
        \tabrowrule
        Right-Turn Yield & Cross traffic from the left traverses the ego vehicle's right-turn path. & 10~m & 10~m & 8--12~m/s; source interval 10--50~m, extended to 60~m on designated routes; one evaluated passage & Flow ends after the first actor clears the ego route (or a 12-s safety limit). Route B keeps a stationary context actor. \\
        \hline\hline
    \end{tabular}
    \caption{Temporal and behavioral specification of the six LANTERN
    scenario families.}
    \label{tab:supp_scenario_specifications}
\end{table}
\FloatBarrier

The warning and event points use 1.5-m spatial tolerances except Vehicle
Cut-In, whose event tolerance is 0.5~m. For the first five families, the
anti-deadlock branch changes only when the external actor begins its scripted
motion; it never overrides ego control. Wrong-way and cross-traffic actors also
have finite safety limits to prevent simulator faults from holding a route
indefinitely, but reaching such a limit is recorded separately from normal
clearance. Recovery is evaluated only after the hazard-clear transition. These
event records distinguish a policy failure from actor spawning, event
activation, or simulator failure in the released per-route results.

\section{Complete Cooperative Unified Score}
\label{sec:supp_complete_cus}

\begingroup
\fontsize{9}{10}\selectfont
\setlength{\abovedisplayskip}{2pt plus .5pt minus .5pt}
\setlength{\belowdisplayskip}{2pt plus .5pt minus .5pt}
\setlength{\abovedisplayshortskip}{1pt plus .5pt}
\setlength{\belowdisplayshortskip}{1pt plus .5pt minus .5pt}
\setlength{\jot}{2pt}
\setlength{\textfloatsep}{5pt plus 1pt minus 1pt}
\setlength{\intextsep}{5pt plus 1pt minus 1pt}
\captionsetup[table]{skip=3pt}

\noindent\textbf{Scoring unit.}
The scoring unit is one matched Route A/B pair. Route A contains the controlled
hazard; Route B retains the route, traffic context, weather, and navigation
objective but disables the target behavior and warning. Scores use the model's
own closed-loop executions, not an expert trajectory. Normalized progress is
$P_i^A=\mathrm{RC}_i^A/100$ and $P_i^B=\mathrm{RC}_i^B/100$, where
$\mathrm{RC}$ is CARLA route completion. All continuous terms are clipped to
$[0,1]$.

\noindent\textbf{Time-to-collision.}
Time-to-Collision (TTC) uses only the controlled actor from the warning
reference until hazard clearance. Let $\mathbf r(t)$ point from ego to the
actor and $d_{\mathrm{box}}(t)$ be their 3D bounding-box surface distance:
\begin{equation}
\begin{aligned}
v_{\mathrm{close}}(t)&=
-\frac{\mathbf r(t)^\mathsf{T}
(\mathbf v_{\mathrm{hazard}}(t)-\mathbf v_{\mathrm{ego}}(t))}
{\lVert\mathbf r(t)\rVert_2},\\
\mathrm{TTC}(t)&=\frac{d_{\mathrm{box}}(t)}{v_{\mathrm{close}}(t)},\qquad
S_{\mathrm{TTC},i}^A=
\operatorname{clip}_{[0,1]}\!\left(\frac{\mathrm{TTC}_{\min,i}}{3~\mathrm{s}}\right).
\end{aligned}
\end{equation}
TTC samples require $v_{\mathrm{close}}>0.05$~m/s. No closing trend scores one,
box overlap scores zero, and evaluator collisions additionally fail a hard
gate.

\noindent\textbf{Clearance.}
From event onset to clearance, let $d_{\min,i}$ be the minimum controlled-actor
surface distance. Then
$S_{\mathrm{clear},i}^A=\operatorname{clip}_{[0,1]}
(d_{\min,i}/d_{\mathrm{safe}})$, with $d_{\mathrm{safe}}=2$~m for Pedestrian
Emergence and Obstacle Reveal and $3$~m otherwise. Collision sets
$d_{\min}=0$ and fails the collision gate.

\noindent\textbf{Anticipation.}
Anticipation (Ant.) locates the first sustained response within the pre-event
interval:
\begin{equation}
S_{\mathrm{ant},i}^A
=\operatorname{clip}_{[0,1]}
\left(\frac{t_{\mathrm{event}}-t_{\mathrm{react}}}
{t_{\mathrm{event}}-t_{\mathrm{ref}}}\right).
\end{equation}
Here $t_{\mathrm{ref}}$ is warning onset for a \textit{warning} execution and
the paired route-defined time otherwise. $t_{\mathrm{react}}$ is the first
0.5-s interval in which at least 80\% of samples have acceleration
$\leq-0.5$~m/s$^2$ or speed at least 1~m/s below the pre-warning reference.
Response at warning onset scores one; response at or after event onset, or no
response, scores zero. With coincident warning/event onset in Right-Turn
Yield, a response must already be present.

\noindent\textbf{Recovery.}
Recovery (Rec.) starts after both warning termination and hazard clearance.
Recovery requires speed above
$\max(2~\mathrm{m/s},0.5v_{\mathrm{pre}})$ for 1~s and at least 0.5~m of
additional progress. For latency
$\mathrm{RL}_i=t_{\mathrm{recover}}-t_{\mathrm{release}}$,
$S_{\mathrm{rec},i}^A=\operatorname{clip}_{[0,1]}
(1-\mathrm{RL}_i/8~\mathrm{s})$. No recovery within 8~s scores zero; blocking
or timeout also fails a hard gate.

\noindent\textbf{Hard gates.}
Route A must pass five binary gates:
\begin{equation}
G_i^A=
G_{\mathrm{coll},i}^A
\,G_{\mathrm{road},i}^A
\,G_{\mathrm{lane},i}^A
\,G_{\mathrm{rule},i}^A
\,G_{\mathrm{block},i}^A.
\end{equation}
The gates fail upon any collision, severe route departure,
\texttt{outside\_route\_lanes} infraction, traffic-rule violation, blocking,
or timeout. Any such failure makes $G_i^A=0$.

If a hard-gate failure occurs before valid hazard telemetry is established,
the corresponding TTC, clearance, anticipation, and recovery fields are left
unavailable in the route-level file. The failed gate still determines
$\mathrm{CUS}_i=0$; an unavailable component is never imputed as a successful
response.

Route B contributes one additional False Brake gate
$G_{\mathrm{FB},i}^B$. It fails only when the ego remains at or below
0.1~m/s for at least 1~s in the interval matched to Route A's warning-to-event
approach, without a traffic signal, stop sign, lead vehicle, obstacle,
pedestrian, intersection conflict, or route end. Other braking does not fail
the gate; other Route B infractions affect CUS only through progress.

\noindent\textbf{Pair score and aggregation.}
\begin{gather}
S_i^A=(S_{\mathrm{TTC},i}^A+S_{\mathrm{clear},i}^A)/2,
\qquad
T_i^A=(S_{\mathrm{ant},i}^A+S_{\mathrm{rec},i}^A)/2,\notag\\
Q_i^A=0.40P_i^A+0.35S_i^A+0.25T_i^A,\notag\\
\boxed{\mathrm{CUS}_i
=100\,G_i^A G_{\mathrm{FB},i}^B
\left(0.85Q_i^A+0.15P_i^B\right)}.
\label{eq:supp_cus_complete}
\end{gather}
Route A supplies 85\% of the continuous score and Route B progress 15\%;
failed hard gates set the pair score to zero. Each family score averages its
20 pairs, and Overall CUS macro-averages the six family scores.

\noindent\textbf{Worked example.}
Table~\ref{tab:supp_cus_example} uses the final
Coop-SimLingo \textit{warning} execution on
\texttt{closed\_occluded\_pedestrian\_0001}. Here $P^A=P^B=1$,
$\mathrm{TTC}_{\min}=2.082$~s, $d_{\min}=10.794$~m,
$(t_{\mathrm{ref}},t_{\mathrm{event}},t_{\mathrm{react}})
=(3.30,5.30,3.35)$~s, $\mathrm{RL}=0.60$~s, and both gate products equal one.

\begin{center}
    \centering
    \setlength{\tabcolsep}{8pt}
    \renewcommand{\arraystretch}{0.98}
    \begin{tabular}{cccc}
        \hline\hline
        Route A progress & TTC & Clearance & Anticipation \\
        \hline\hline
        $1.000\rightarrow0.34000$ &
        $0.694\rightarrow0.10323$ &
        $1.000\rightarrow0.14875$ &
        $0.975\rightarrow0.10359$ \\
        \tabrowrule
        Recovery & Route B progress & Gate product & Final CUS \\
        \tabrowrule
        $0.925\rightarrow0.09828$ &
        $1.000\rightarrow0.15000$ &
        $1\times1$ &
        $\mathbf{100(0.94386)=94.39}$ \\
        \hline\hline
    \end{tabular}
    \captionof{table}{Worked CUS example; arrows map normalized scores to weighted contributions.}
    \label{tab:supp_cus_example}
\end{center}
The six contributions sum to $0.94386$; any failed gate would instead make the
pair score zero.
\FloatBarrier
\endgroup

\section{Protocol Extension to Unseen Hazards}

\noindent\textbf{Controlled case study.}
We construct an occluded-cyclist crossing that is absent from the six LANTERN
training families. The route is an unused 129.1-m Town15 segment sourced from
\texttt{DynamicObjectCrossing}; it does not occur in the frozen 120-pair test
manifest. A cyclist travels at 4.5~m/s from behind a parked bus and crosses the
ego lane. The route otherwise retains the warning-on, event-on, direct-
observation, and event-clear transitions of the main protocol. We evaluate the
released Coop-SimLingo checkpoint without additional fine-tuning under three
conditions: \textit{no-warning}; the free-form warning ``A cyclist hidden
behind the parked van at \texttt{<FRONT\_CAR\_POS>} is about to cross the ego
lane''; and the held-out paraphrase ``Watch for a bicycle rider emerging from
behind the roadside vehicle near \texttt{<FRONT\_CAR\_POS>}; the rider will
enter your path.'' As in the main benchmark, the token shown here denotes a
template slot and is replaced online by the grounded two-dimensional hazard
location. Route geometry, weather, background-traffic seed, checkpoint,
and physical-event logic are held fixed.

\begin{figure}[htbp]
    \centering
    \includegraphics[width=0.30\textwidth]{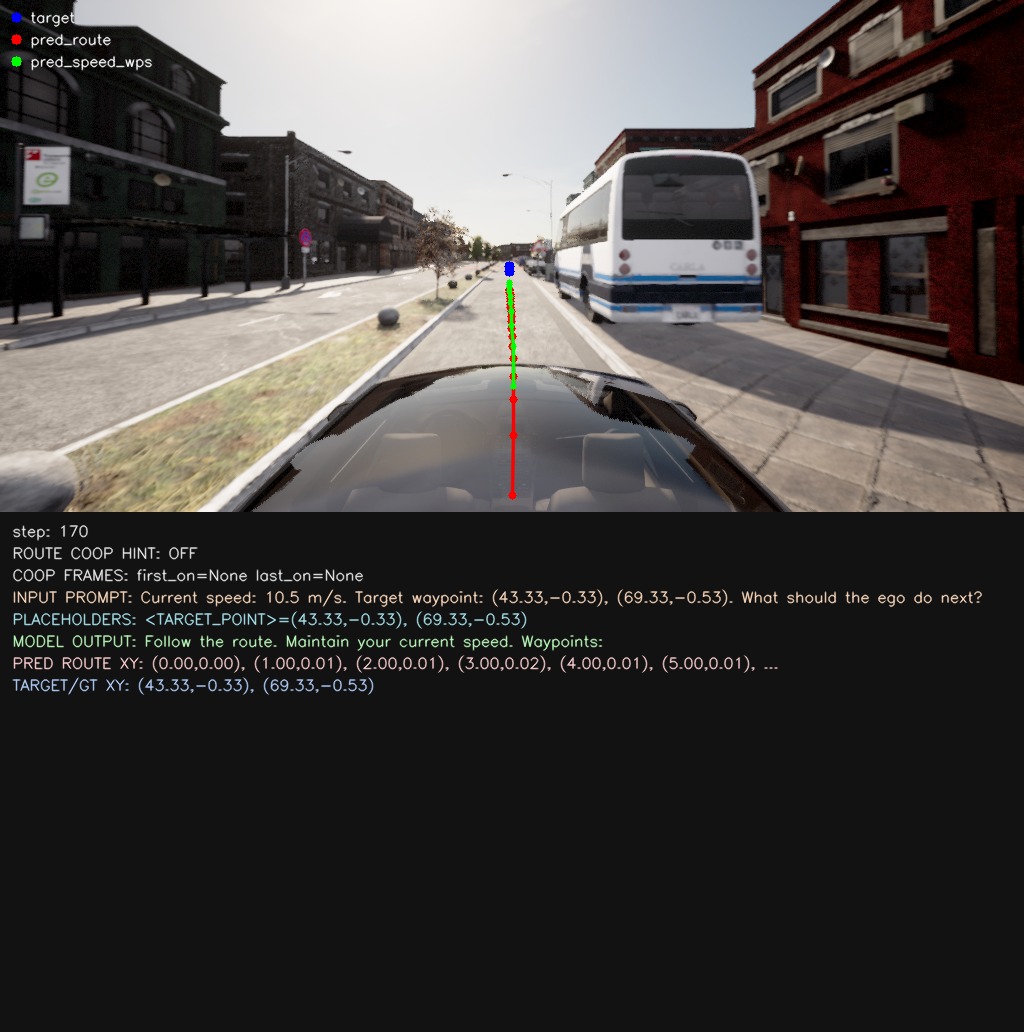}\hfill
    \includegraphics[width=0.30\textwidth]{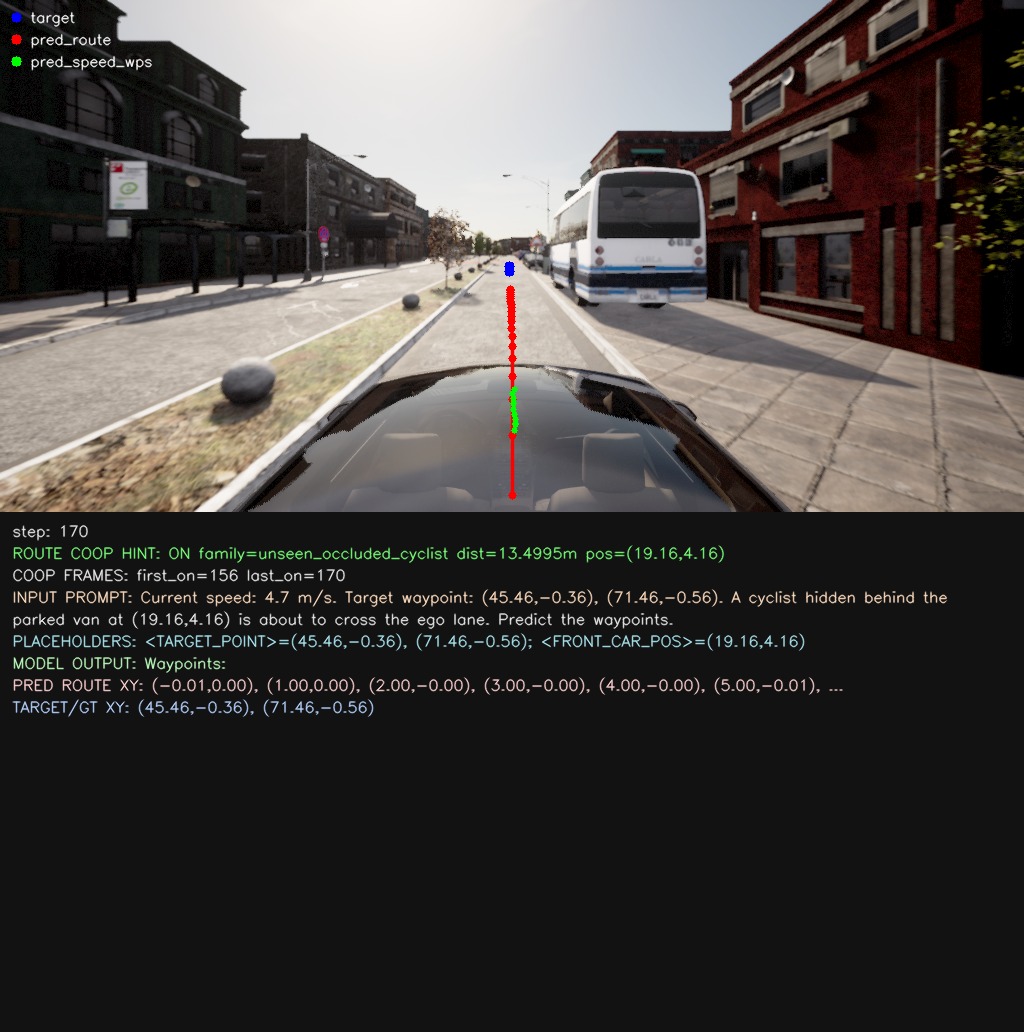}\hfill
    \includegraphics[width=0.30\textwidth]{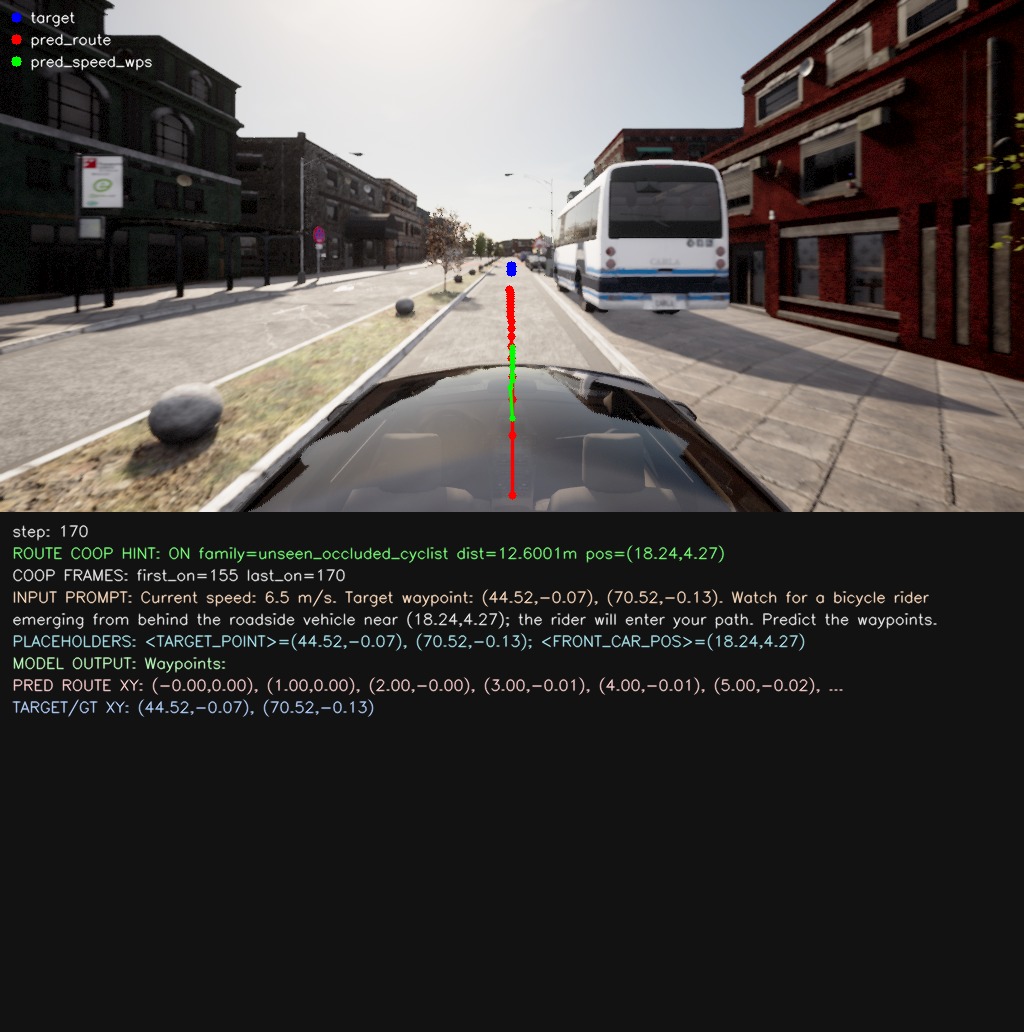}
    \caption{Matched predictions for the unseen cyclist event at simulation
    step 170: no warning (left), free-form warning (center), and held-out
    paraphrase (right). At this point, the ego speed is 10.5, 4.7, and
    6.5~m/s, respectively.}
    \label{fig:supp_unseen_cyclist}
\end{figure}

We define the response time as the first of three consecutive 20-Hz control
steps with brake at least 0.5. Response lead is event-on time minus response
time, so a positive value indicates braking before the cyclist begins to
cross. This case-study response lead is separate from the CUS Anticipation
score, which uses the sustained-response definition in
Section~\ref{sec:supp_complete_cus}. Pre-event speed reduction is measured from warning onset to the minimum
speed before event-on. Table~\ref{tab:supp_unseen_cyclist} shows that both
unseen wordings cause a sustained response before event-on, whereas the
\textit{no-warning} response begins only after the event. All three executions
ultimately complete without collision or blocking; the distinction is therefore
one of anticipatory timing rather than eventual route success. Figure~\ref{fig:supp_unseen_cyclist}
shows the corresponding predictions at the same simulation step.

\begin{center}
    \refstepcounter{table}\label{tab:supp_unseen_cyclist}
    \small
    \setlength{\tabcolsep}{7pt}
    \renewcommand{\arraystretch}{1.12}
    \begin{tabular}{lrrrrr}
        \hline\hline
        Condition & Lead (s) $\uparrow$ & $\Delta v$ (m/s) & RC & Coll. & Block \\
        \hline\hline
        \textit{no-warning}       & $-0.60$ & 0.0 & 100 & 0 & 0 \\
        \tabrowrule
        Free-form warning& 1.95    & 9.5 & 100 & 0 & 0 \\
        \tabrowrule
        Held-out paraphrase & 1.80 & 9.3 & 100 & 0 & 0 \\
        \hline\hline
    \end{tabular}
    \par\vspace{2pt}
    \begin{minipage}{0.96\columnwidth}
        \footnotesize\textbf{Table \thetable:} Held-out occluded-cyclist case
        study. RC, collision, and blocking are percentages; $\Delta v$ is the
        pre-event speed reduction.
    \end{minipage}
\end{center}

\noindent\textbf{Additional physical events.}
We further construct two route-disjoint events whose actor semantics are absent
from the six LANTERN families: an ambulance that crosses the ego path from
behind a large vehicle at a Town15 intersection, and a parked vehicle that
reverses into the ego lane on a Town13 road. Both routes are reconstructed from
recorded Bench2Drive trajectories that are absent from the frozen test manifest.
We again fix the route, weather, traffic seed, checkpoint, and actor behavior
across the three input conditions and perform no additional fine-tuning.

The exact warning transfers successfully in both cases
(Table~\ref{tab:supp_additional_unseen}). For the emergency crossing, it moves
the sustained response from 0.55~s after event onset to 2.00~s before onset,
eliminates the collision, and raises route completion from 55.1 to 100. For the
reversing vehicle, all executions eventually complete, but only the exact
warning causes a sustained pre-event response.

\begin{center}
    \refstepcounter{table}\label{tab:supp_additional_unseen}
    \small
    \setlength{\tabcolsep}{5pt}
    \renewcommand{\arraystretch}{1.12}
    \begin{tabular}{llrrrrr}
        \hline\hline
        Event & Condition & Lead (s) $\uparrow$ & $\Delta v$ (m/s) & RC & Coll. & Block \\
        \hline\hline
        Emergency crossing & \textit{no-warning} & $-0.55$ & 0.0 & 55.1 & 1 & 0 \\
        \tabrowrule
        & Exact warning & 2.00 & 8.9 & 100 & 0 & 0 \\
        \tabrowrule
        & Held-out paraphrase & 1.35 & 3.5 & 55.1 & 1 & 0 \\
        \tabrowrule
        Reversing vehicle & \textit{no-warning} & $-0.45$ & 1.8 & 100 & 0 & 0 \\
        \tabrowrule
        & Exact warning & 1.95 & 10.6 & 100 & 0 & 0 \\
        \tabrowrule
        & Held-out paraphrase & $-0.60$ & 0.0 & 100 & 0 & 0 \\
        \hline\hline
    \end{tabular}
    \par\vspace{2pt}
    \begin{minipage}{0.96\columnwidth}
        \footnotesize\textbf{Table \thetable:} Controlled transfer to two
        additional physical events. Lead and $\Delta v$ follow the definitions
        in Table~\ref{tab:supp_unseen_cyclist}; collision and blocking are
        route-level counts.
    \end{minipage}
\end{center}

\begin{figure}[htbp]
    \centering
    \includegraphics[width=0.44\textwidth]{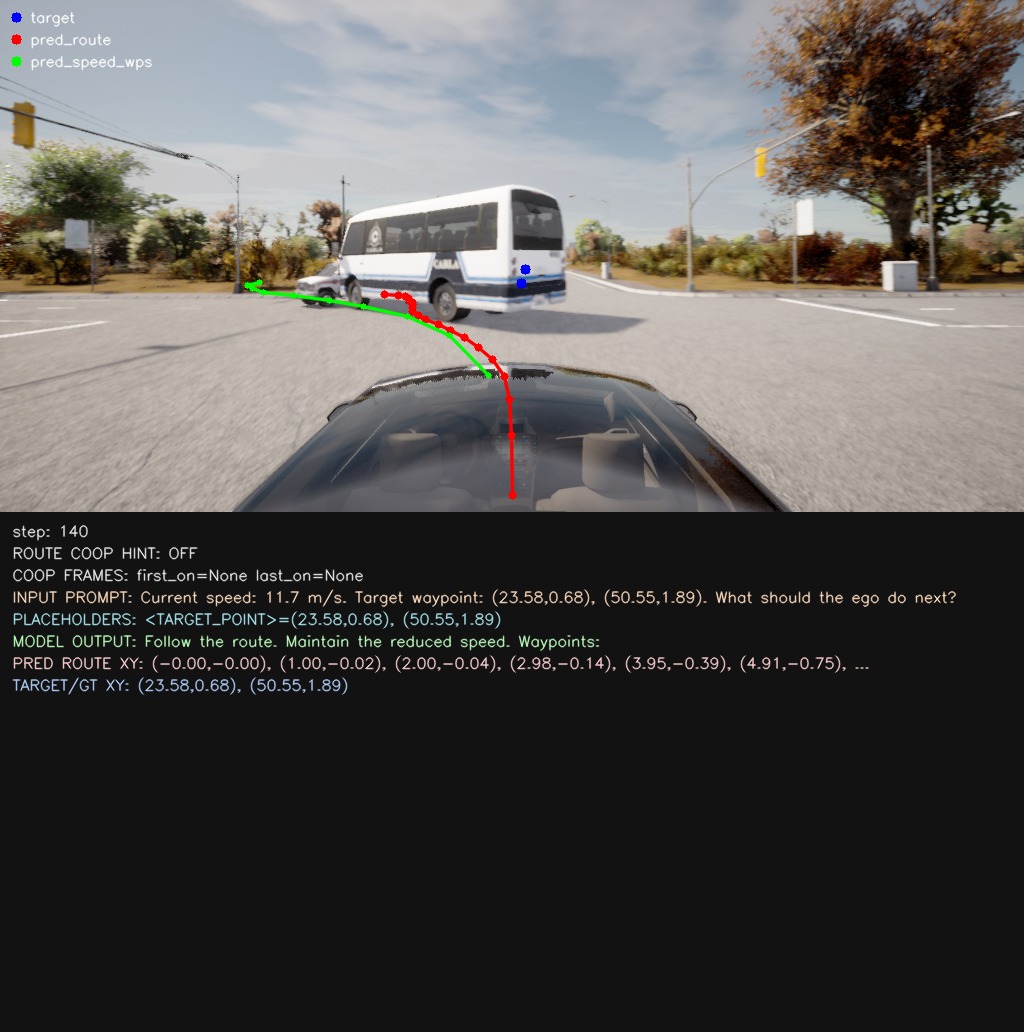}\hfill
    \includegraphics[width=0.44\textwidth]{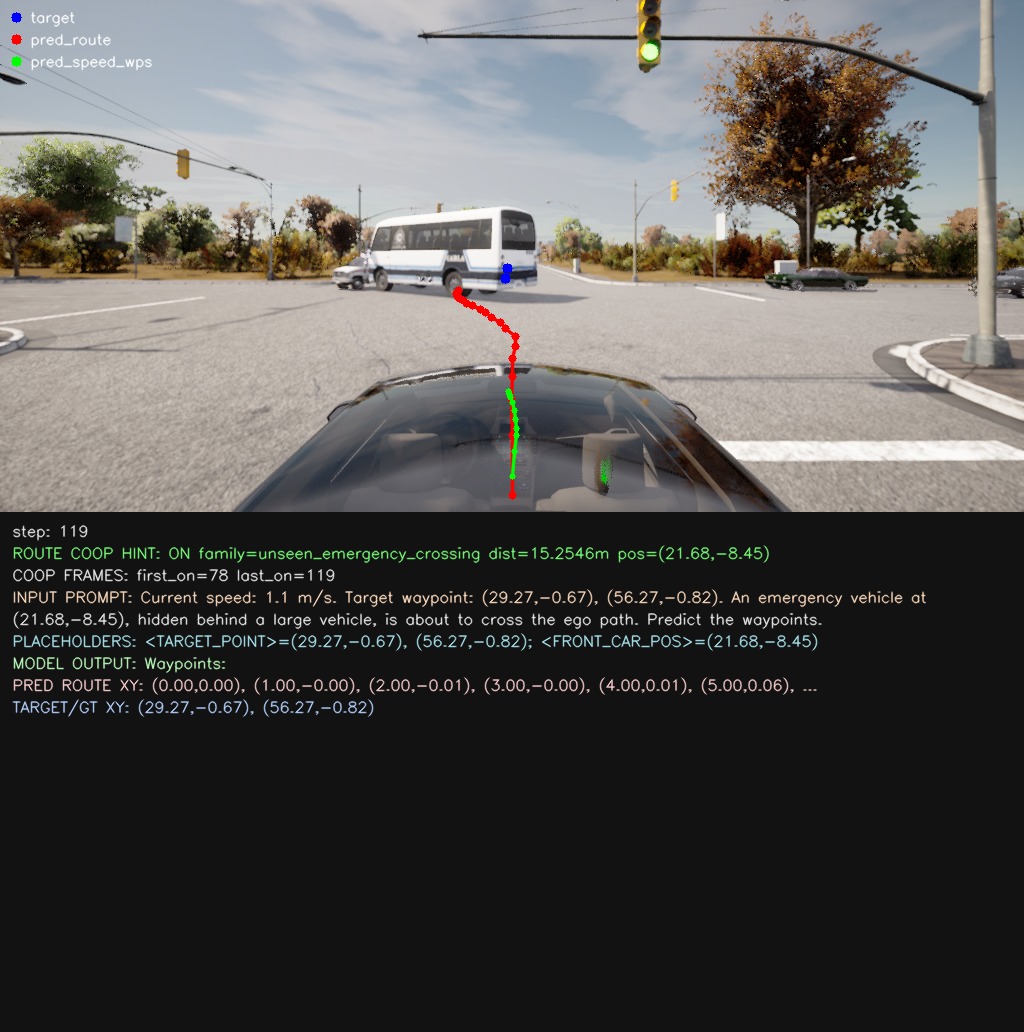}

    \vspace{3pt}
    \includegraphics[width=0.44\textwidth]{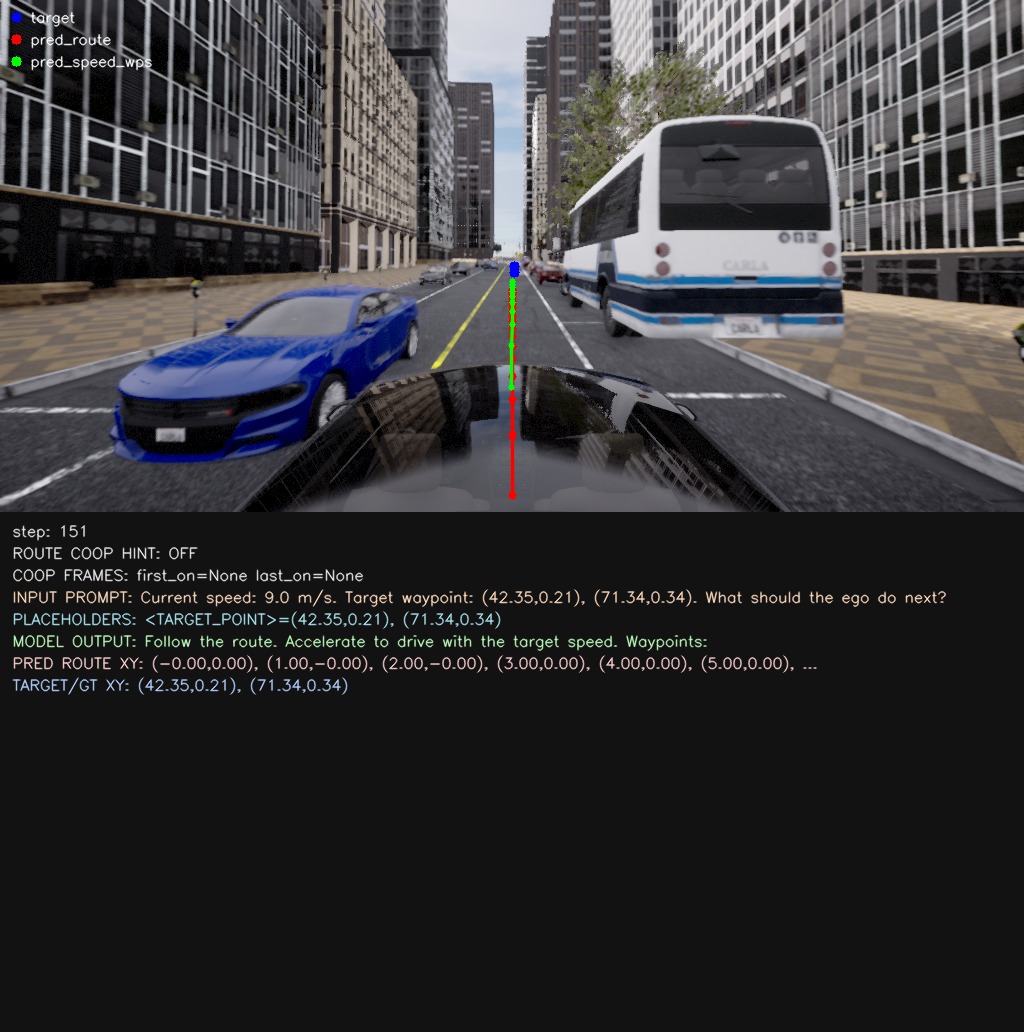}\hfill
    \includegraphics[width=0.44\textwidth]{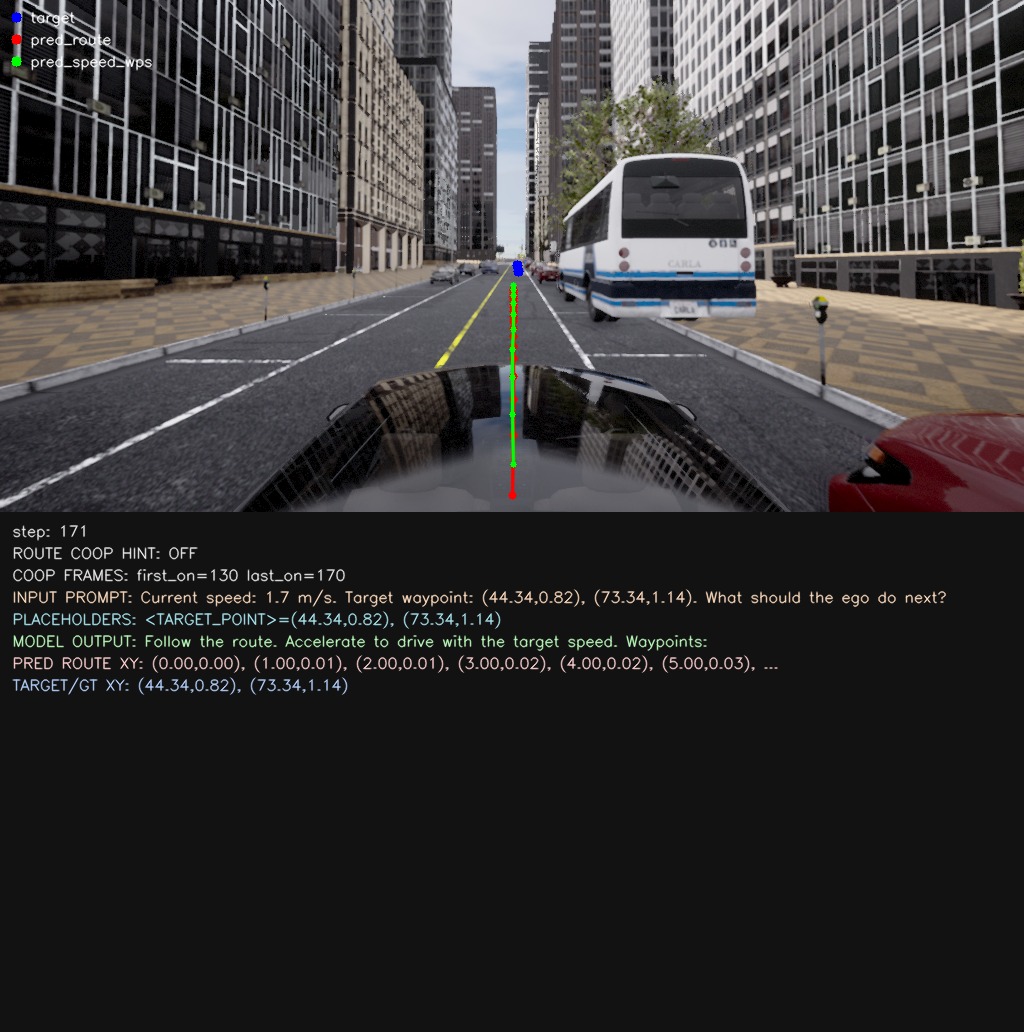}
    \caption{Matched event-on predictions for the emergency crossing (top) and
    reversing vehicle (bottom). Left panels are from \textit{no-warning}
    rollouts; right panels are from rollouts that received the exact warning,
    which may already have terminated by the displayed event-on frame. The
    warning-conditioned rollouts exhibit substantially reduced predicted
    progress.}
    \label{fig:supp_additional_unseen}
\end{figure}

Together, the cyclist, emergency-vehicle, and reversing-vehicle examples show
that LANTERN's temporal protocol can be instantiated for new actor behaviors
while preserving matched event timing and closed-loop evaluation. Across all
three events, an accurate free-form warning produces an anticipatory response
before the new hazard begins.

\section{Additional Ablation Analysis}

\noindent\textbf{Localization noise.}
The main-paper ablation reports the same one-decimal CUS, 75.5, for the full
protocol and for two-meter warning-position noise. This equality does not imply
identical behavior. Under localization noise, Route A progress rises from 95.4
to 98.4 and blocking falls from 5.8 to 3.3, while collisions increase from 0.0
to 3.4 and recovery changes from 82.4 to 82.1. These offsetting effects produce
the same rounded aggregate. The perturbation leaves the warning semantics and
timing unchanged and usually keeps the reported position within the same local
interaction region. The result therefore indicates tolerance to moderate
localization error, not invariance to warning location.

\noindent\textbf{Warning wording.}

The held-out warning-paraphrase result reported in the main paper aggregates
three levels of linguistic ambiguity. We disaggregate it here to distinguish
sensitivity to surface wording from sensitivity to the event and location
information carried by the warning. The 120 closed-loop routes are partitioned
into three disjoint 40-route subsets. Low ambiguity changes local wording while
preserving the event and location; medium ambiguity describes the event less
directly; high ambiguity removes the location and retains only a generic
description of the impending interaction. Each valid paraphrase execution is
compared with the canonical-warning execution of the same route and checkpoint.
One low-, one medium-, and two high-ambiguity executions lack complete
infrastructure records and are excluded from both sides of the corresponding
comparison.

\begin{table}[htbp]
    \centering
    \small
    \setlength{\tabcolsep}{4pt}
    \renewcommand{\arraystretch}{1.12}
    \begin{tabular*}{\columnwidth}{@{\extracolsep{\fill}}lrrrr@{}}
        \hline\hline
        Wording & Paraphrase CUS & Matched canonical & Retention & Coll. \\
        \hline\hline
        Low ambiguity ($n=39$) & 83.1 & 77.9 & 106.7\% & 0.0\% \\
        \tabrowrule
        Medium ambiguity ($n=39$) & 51.1 & 69.2 & 73.9\% & 28.2\% \\
        \tabrowrule
        High ambiguity, no position ($n=38$) & 37.5 & 69.3 & 54.1\% & 50.0\% \\
        \hline\hline
    \end{tabular*}
    \caption{Warning-wording ablation by ambiguity. Retention is paraphrase CUS
    divided by canonical-warning CUS on the same valid routes.}
    \label{tab:supp_language_generalization}
\end{table}

Because the ambiguity levels use disjoint route subsets, we interpret each row
against its matched canonical baseline rather than comparing raw CUS across
rows. Low-ambiguity wording preserves performance and incurs no collisions,
showing that the model is not tied to an exact sentence template. Medium
ambiguity reduces matched CUS by 18.1 points, while removing both explicit event
detail and location reduces it by 31.8 points and raises collisions to 50.0\%.
The degradation therefore arises primarily when a warning loses actionable
semantic or spatial content, rather than from minor lexical variation alone.

\begin{figure}[htbp]
    \centering
    \includegraphics[width=0.44\columnwidth]{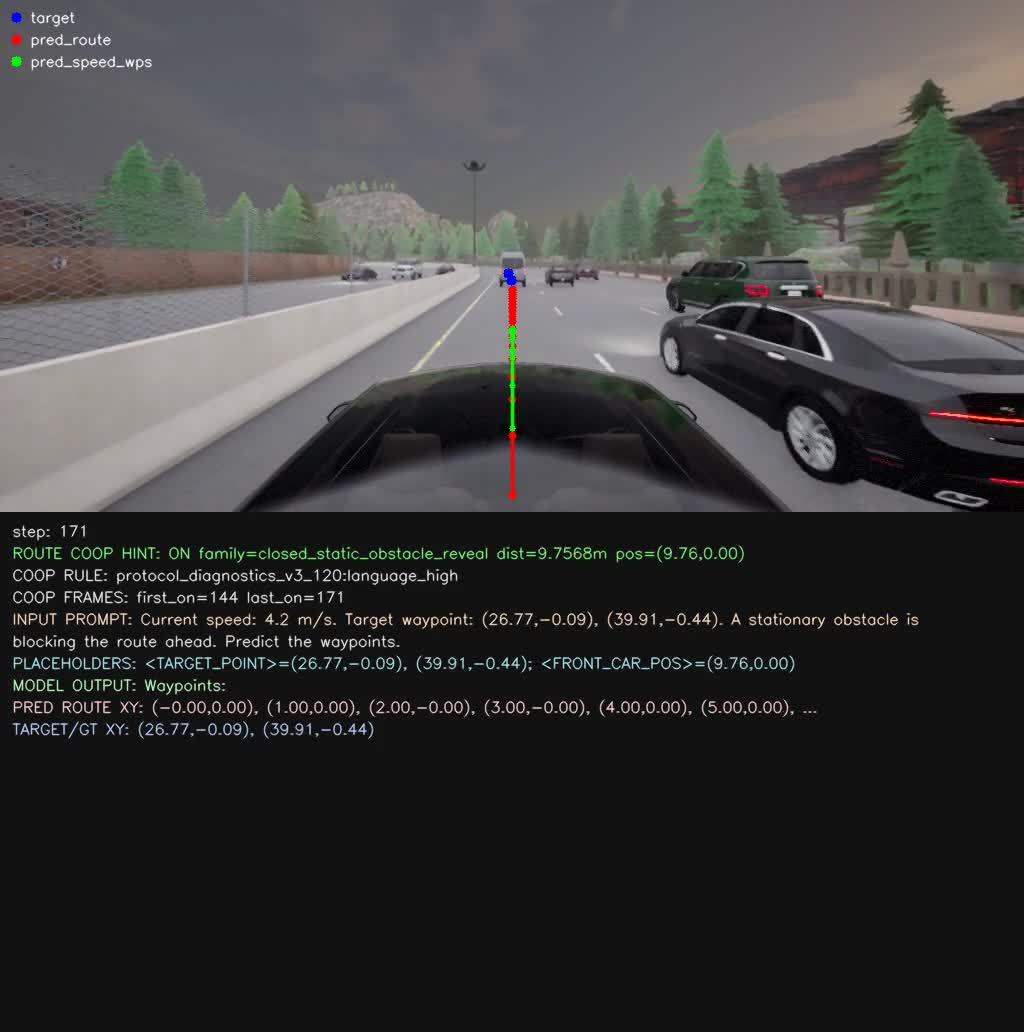}\hfill
    \includegraphics[width=0.44\columnwidth]{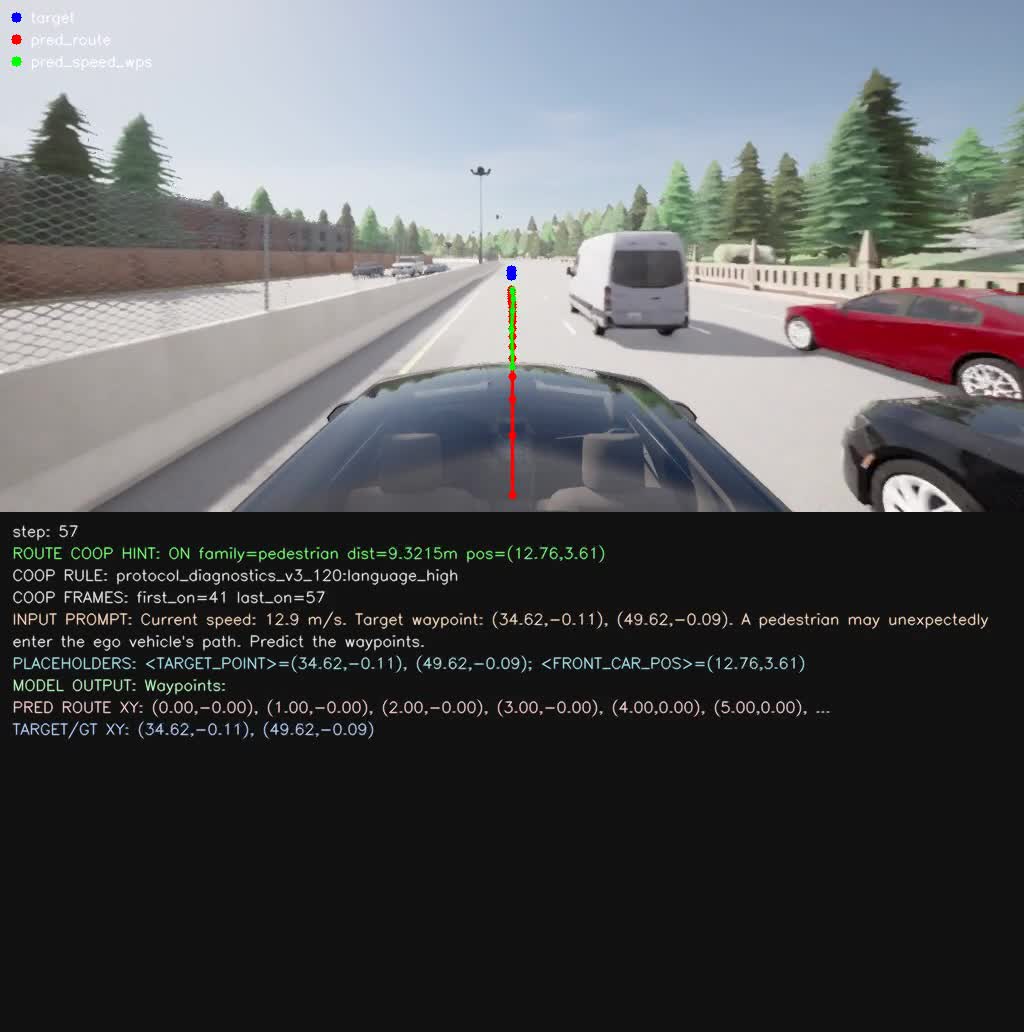}
    \caption{Representative high-ambiguity, position-free warnings. The
    Obstacle Reveal route completes successfully (left), whereas the Pedestrian
    Emergence route collides after an insufficient response (right).}
    \label{fig:supp_language_generalization}
\end{figure}

\FloatBarrier
\section{Diagnostics with Non-Actionable Language Inputs}

\noindent\textbf{Distance-conditioned warnings.}
This diagnostic uses a separate checkpoint fine-tuned only on Pedestrian
Emergence; it is not the Coop-SimLingo checkpoint evaluated in the main
benchmark. We test whether explicit counterexamples can teach a driving model
to reject a warning whose reported geometry is incompatible with an imminent
local hazard. The diagnostic checkpoint pairs a nearby
warning with a stopping trajectory while retaining the original expert
trajectory for the same warning at a sufficiently distant location. We
evaluate three held-out Route B frames under \textit{no-warning}, a warning at
$(12.0,0.5)$~m, and the same warning at $(100.0,0.5)$~m.

\begin{figure}[H]
    \centering
    \includegraphics[width=\textwidth]{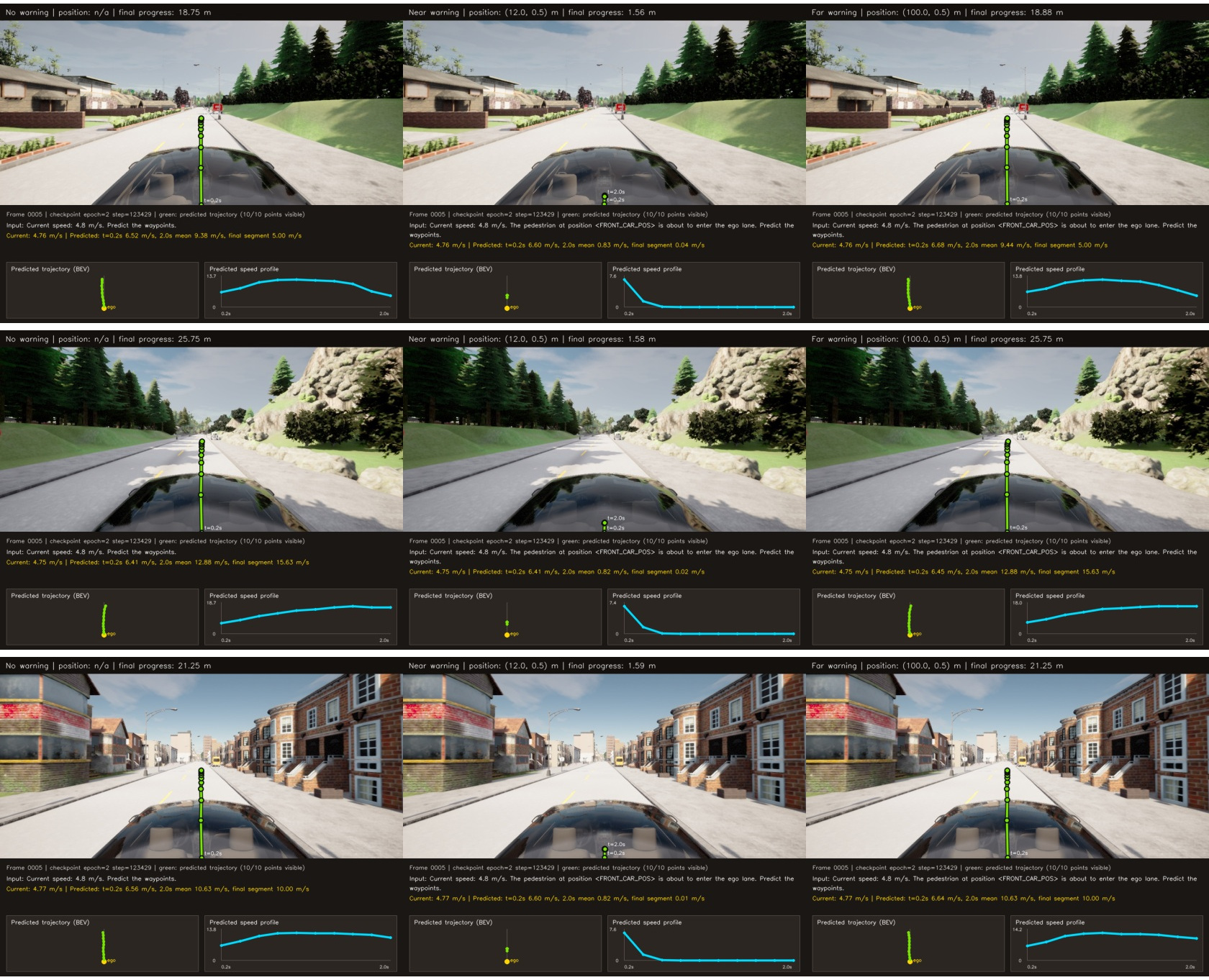}
    \caption{Distance-conditioned responses on three held-out S1 control
    frames. Columns show \textit{no-warning}, a nearby warning, and the same
    warning at 100~m. The nearby warning produces a stopping trajectory,
    whereas the distant warning preserves ordinary progress.}
    \label{fig:supp_far_warning_rejection}
\end{figure}

\begin{center}
    \refstepcounter{table}\label{tab:supp_far_warning_rejection}
    \small
    \setlength{\tabcolsep}{7pt}
    \renewcommand{\arraystretch}{1.12}
    \begin{tabular}{lrrrr}
        \hline\hline
        Sample & \textit{no-warning} & Near & Far & Retention \\
        \hline\hline
        Route 1 & 18.75 & 1.56 & 18.88 & 100.7\% \\
        \tabrowrule
        Route 2 & 25.75 & 1.58 & 25.75 & 100.0\% \\
        \tabrowrule
        Route 3 & 21.25 & 1.59 & 21.25 & 100.0\% \\
        \hline\hline
    \end{tabular}
    \par\vspace{2pt}
    \begin{minipage}{0.96\columnwidth}
        \footnotesize\textbf{Table \thetable:} Final predicted progress in
        meters for a checkpoint trained with distance-conditioned S1
        counterexamples. Retention is far-warning progress divided by
        \textit{no-warning} progress.
    \end{minipage}
\end{center}

Table~\ref{tab:supp_far_warning_rejection} and
Figure~\ref{fig:supp_far_warning_rejection} show that the model need not react
unconditionally to every hazard statement: within the trained warning family,
it uses the reported distance to separate an actionable warning from a distant
one. This demonstrates learned selectivity, not general contradiction
reasoning. A narrow negative set can teach a template-specific rule without
covering qualitatively different inconsistencies, such as a hazard behind the
ego vehicle, an impossible road configuration, or an unseen warning type.
Robust rejection of such cases would require counterexamples that vary jointly
over hazard semantics, location, road context, and wording. The experiment
therefore demonstrates the value of targeted negative supervision while
leaving broader visual--language consistency reasoning as an open problem.

\noindent\textbf{Non-actionable warnings.}
We next test warnings that describe nearby activity but explicitly place it
outside the ego vehicle's driving path. None duplicates a warning used by
LANTERN. All five inputs are evaluated on the same empty-road route. A warning
is treated as non-actionable when, relative to the \textit{no-warning}
execution, its maximum additional speed reduction is below 2~m/s, it causes no
sustained stop of at least 1~s, and route completion decreases by no more than
five percentage points.

\begin{center}
    \refstepcounter{table}\label{tab:supp_irrelevant_text}
    \small
    \setlength{\tabcolsep}{6pt}
    \renewcommand{\arraystretch}{1.10}
    \begin{tabular}{p{0.53\textwidth}rrr}
        \hline\hline
        Warning input & $\Delta v_{\max}$ & Stop & RC \\
        \hline\hline
        Caution: maintenance equipment is operating behind the roadside
        barrier. & 1.4 & 0.0 & 100 \\
        \tabrowrule
        Warning: a delivery van is parked in an off-road service area. &
        2.8 & 0.0 & 100 \\
        \tabrowrule
        Caution: a pedestrian is waiting behind a roadside fence and remains
        clear of traffic. & 1.9 & 0.0 & 100 \\
        \tabrowrule
        Warning: a cyclist is traveling away on a physically separated path.
        & 0.4 & 0.0 & 100 \\
        \tabrowrule
        Caution: traffic is restricted on a parallel road outside the ego
        route. & 1.8 & 0.0 & 100 \\
        \hline\hline
    \end{tabular}
    \par\vspace{2pt}
    \begin{minipage}{0.96\columnwidth}
        \footnotesize\textbf{Table \thetable:} Response to non-actionable
        warnings. $\Delta v_{\max}$ is the maximum additional speed reduction
        in m/s, Stop is the longest stationary duration in seconds, and RC is
        route completion in percent.
    \end{minipage}
\end{center}

\begin{figure}[H]
    \centering
    \includegraphics[width=0.80\textwidth]{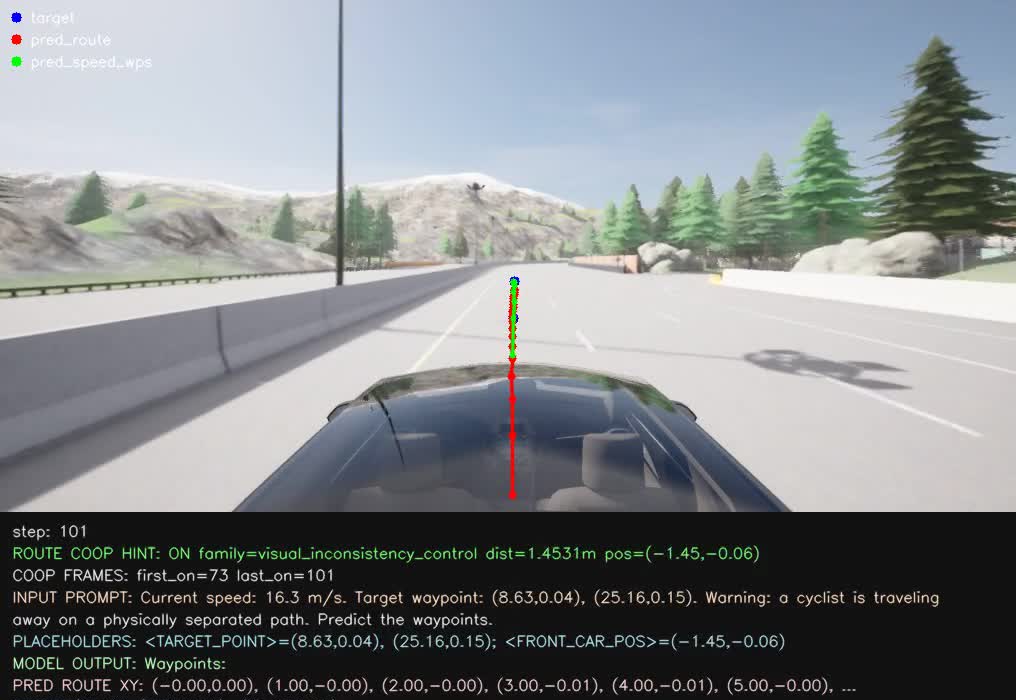}
    \caption{Representative response to a warning about a cyclist traveling
    away on a physically separated path. The predicted trajectory continues
    along the route without sustained stopping.}
    \label{fig:supp_irrelevant_text}
\end{figure}

Four of the five warnings satisfy the strict non-actionable criterion. All
five complete the route without collision, blocking, or sustained stopping,
and their mean maximum additional speed reduction is 1.66~m/s. The warning
about a parked delivery van causes a transient 2.8~m/s reduction, showing that
actor-related wording can still induce conservative behavior. Together with
the distance-conditioned result, this test shows that the model does not
unconditionally stop for every warning, while leaving general
visual--language consistency reasoning as an open problem.
\FloatBarrier

\section{Complete Results and Reproducibility}

The released \texttt{summary/reproducibility/route\_level\_results.csv}
indexes all 720 final model--route pairs with Route A/B metrics, gate values,
artifact provenance, and optional video paths. Its companion schema defines
all fields, units, and missing-value rules. In particular, continuous hazard
terms remain empty when a hard-gate failure occurs before valid hazard
telemetry; they are never imputed as successful responses. Audit files retain
infrastructure failures and accepted replacements without converting one into
the other.

The machine-readable \texttt{reproducibility\_manifest.json} records CARLA
0.9.15, Python 3.8.18, PyTorch 2.2.0 with CUDA 12.1, Transformers 4.46.3,
Traffic Manager seed 0, checkpoint identifiers and hashes, model adapters,
portable validation and route commands, expected outputs, and aggregate metric
provenance. The frozen 120-pair v1.1 route manifest has SHA-256 prefix
\texttt{6b4389fd91d3}. Together, these files reproduce the reported aggregates
and preserve the provenance of every included execution.

\end{document}